\documentclass[default,iicol, sn-apacite]{sn-jnl}

\usepackage{hyperref}
\usepackage{graphicx}%
\usepackage{multirow}%
\usepackage{amsmath,amssymb,amsfonts}%
\usepackage{amsthm}%
\usepackage{mathrsfs}%
\usepackage[title]{appendix}%
\usepackage{xcolor}%
\usepackage{textcomp}%
\usepackage{manyfoot}%
\usepackage{booktabs}%
\usepackage{algorithm}%
\usepackage{algorithmicx}%
\usepackage{algpseudocode}%
\usepackage{listings}%

\usepackage{color, multirow, multicol, verbatim, threeparttable, diagbox, makecell, booktabs, colortbl,setspace}

\usepackage[authoryear,round]{natbib}

\theoremstyle{thmstyleone}%
\theoremstyle{thmstyletwo}%

\theoremstyle{thmstylethree}%

\begin{document}

\title[Article Title]{Pay Less Attention to Deceptive Artifacts: Robust Detection of Compressed Deepfakes on Online Social Networks}


\author[1,2]{\fnm{Manyi} \sur{Li}}

\author*[3]{\fnm{Renshuai} \sur{Tao}}\email{rstao@bjtu.edu.cn}

\author[2]{\fnm{Yufan} \sur{Liu}}

\author[3]{\fnm{Chuangchuang} \sur{Tan}}

\author[3]{\fnm{Haotong} \sur{Qin}}

\author[2]{\fnm{Bing} \sur{Li}}

\author[3]{\fnm{Yunchao} \sur{Wei}}

\author[3]{\fnm{Yao} \sur{Zhao}}

\affil[1]{\orgdiv{School of Artificial Intelligence}, \orgname{University of Chinese Academy of Sciences}. \orgaddress{\city{Beijing}, \postcode{100190}, \country{China}}}

\affil[2]{\orgdiv{State Key Laboratory of Multimodal Artificial Intelligence Systems, Institute of Automation}, \orgname{Chinese Academy of Sciences}, \orgaddress{ \city{Beijing}, \postcode{100190}, \country{China}}}

\affil[3]{\orgdiv{Institute of Information Science}, \orgname{Beijing Jiaotong University}, \orgaddress{\city{Beijing}, \postcode{100044}, \country{China}}}

\abstract{With the rapid advancement of deep learning, particularly through generative adversarial networks (GANs) and diffusion models (DMs), AI-generated images, or ``deepfakes", have become nearly indistinguishable from real ones. These images are widely shared across Online Social Networks (OSNs), raising concerns about their misuse. Existing deepfake detection methods overlook the ``block effects" introduced by compression in OSNs, which obscure deepfake artifacts, and primarily focus on raw images, rarely encountered in real-world scenarios. To address these challenges, we propose PLADA (Pay Less Attention to Deceptive Artifacts), a novel framework designed to tackle the lack of paired data and the ineffective use of compressed images. PLADA consists of two core modules: Block Effect Eraser (B2E), which uses a dual-stage attention mechanism to handle block effects, and Open Data Aggregation (ODA), which processes both paired and unpaired data to improve detection. Extensive experiments across 26 datasets demonstrate that PLADA achieves a remarkable balance in deepfake detection, outperforming SoTA methods in detecting deepfakes on OSNs, even with limited paired data and compression. More importantly, this work introduces the ``block effect" as a critical factor in deepfake detection, providing a robust solution for open-world scenarios. Our code is available at \url{https://github.com/ManyiLee/PLADA}.}




\maketitle

\begin{figure}[!t]
  \centering
  \setlength{\abovecaptionskip}{-0.1cm}
\includegraphics[scale=0.8]{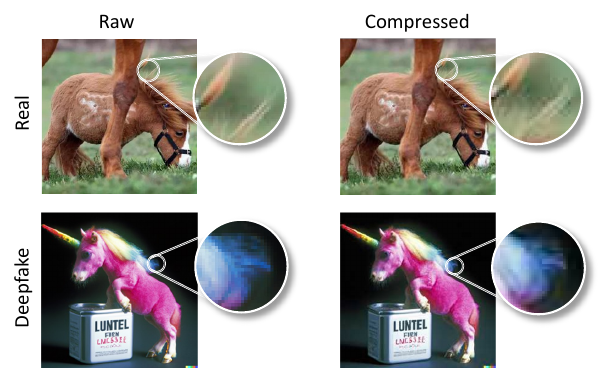}
   \caption{\textbf{Similarity Between the Proposed ``Block Effect" and Deepfake Artifacts.} Compressed images introduce the block effect, which visually resembles deepfake artifacts. This compression can degrade deepfake artifacts, potentially misguiding detectors.}
   \label{block_effect}
\end{figure}

\section{Introduction}
With the rapid advancement of deep learning, particularly the emergence of generative adversarial networks (GANs) \citep{GAN} and diffusion models (DMs) \citep{DM}, AI-generated images have become nearly indistinguishable from real ones \citep{yu2024mining}. These images, often created from noise patterns or text prompts and refined by generative models, have attracted significant academic interest and are commonly referred to as ``deepfakes'' \citep{xu2024fd}. The accessibility of deepfake technology has been greatly enhanced through open-source projects and commercial applications, making it available even to non-experts. As a result, deepfakes are widely disseminated across Online Social Networks (OSNs), raising concerns about their misuse, including the creation of fake news, political scandals, and explicit content.

Despite significant progress in deepfake detection, most existing methods \citep{wang2024watcher, cao2024towards, shao2025robust} focus on raw images, which are rarely encountered in real-world scenarios, particularly on OSNs, where images are often compressed using algorithms like JPEG. This compression introduces block effects that degrade image quality and can alter or eliminate subtle deepfake artifacts, misleading detection models. The block effect, visible in highly compressed images, shares notable similarities with forgery artifacts, as shown in Figure \ref{block_effect}. While some studies \citep{li2023exposingOSNs, qu2024df} have improved detection robustness for OSN-processed images, they primarily focus on general image quality rather than the specific challenges posed by compression. Efforts to enhance robustness against compression, such as mitigating information loss \citep{nguyen2024laa}, have shown limited success in defending against compression attacks. Consequently, researchers \citep{le2023quality, xu2023learningPairwise} have turned to paired data to capture subtle correlations between distributions, but obtaining sufficient paired data remains a significant challenge. However, \textbf{these works fail to simultaneously address two critical problems}: 1) Unpaired data (before and after compression) is often scarce, as obtaining the corresponding raw image is inherently difficult. 2) Compressed images are not effectively utilized, especially when false artifacts caused by block effects are present during compression.

To tackle these challenges, we propose the PLADA (\textbf{P}ay \textbf{L}ess \textbf{A}ttention to \textbf{D}eceptive \textbf{A}rtifacts) method, a novel framework comprising two core modules: Block Effect Eraser (B2E) and Open Data Aggregation (ODA) and. The B2E module introduces a dual-stage attention shift mechanism, incorporating two knowledge injection strategies to guide attention at each backbone layer. The first stage, Residual Guidance (RG), applies an initial attention shift by integrating a multi-head self-attention (MSA) mechanism that fuses information from the Guide Prompt Pool in a residual manner, ensuring minimal interference while providing effective guidance. The second stage, Coordination Guidance (CG), enhances efficiency by integrating global and local features in a bottom-up manner, accelerating attention shifts. Unlike DIKI \citep{DIKI}, which uses a single specialized MSA with prompts, or PKT \citep{PKT}, which combines traditional MSA with a novel variant, B2E introduces a dual-stage attention mechanism featuring two distinct MSA variants. This optimization refines attention shifts at both fine- and coarse-grained levels to better handle compressed deepfakes. The ODA module processes both paired and unpaired data, clustering them into compressed and uncompressed groups. Within each cluster, ODA further calculates the centers of authentic and fake images, maximizing the distance between them. ODA not only promotes distinction between real and fake images but also facilitates interactions between compressed and raw images. By maintaining a strong separation between raw real and raw fake images, ODA significantly constrains the convergence of compressed real and compressed fake images, thus improving detection robustness.

\begin{figure}[!t]
  \centering
  \setlength{\abovecaptionskip}{-0.15cm}
   \includegraphics[scale=0.44]{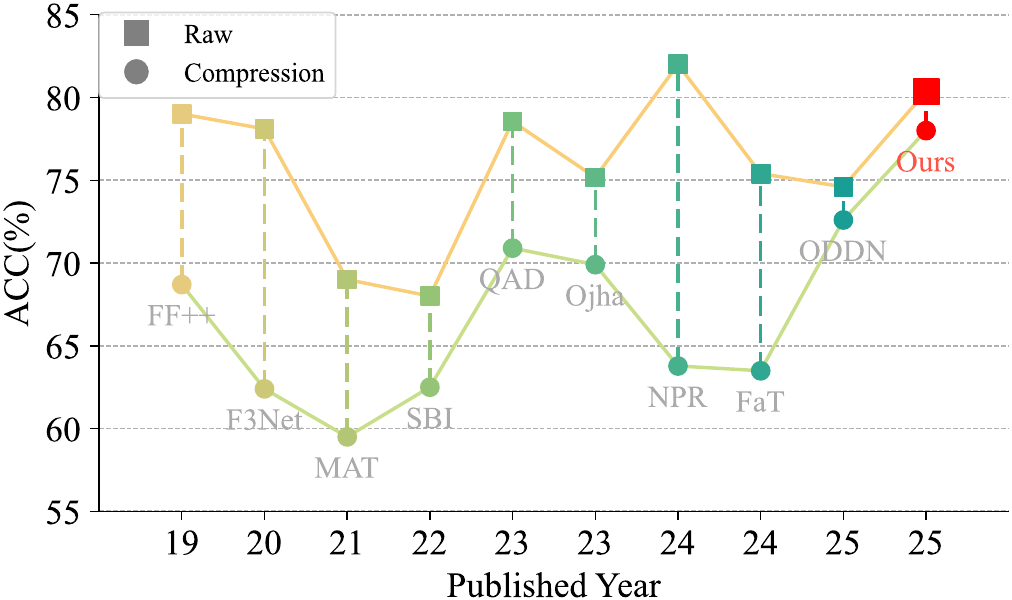}
   \caption{\textbf{Timeline of Techniques and Performance Changes Across Different Image Settings.} PLADA consistently outperforms other approaches, maintaining an optimal balance between raw and compressed images with minimal degradation.}
   \label{introduction}
\end{figure}

Extensive experiments demonstrate that the PLADA method achieves an exceptional balance in deepfake detection accuracy across both uncompressed and compressed images, as shown in Figure \ref{introduction}. Furthermore, it delivers promising performance in detecting compressed deepfakes on OSNs, even with limited paired data, showcasing its potential for handling open-world scenarios. The key contributions are summarized as follows:
\begin{itemize}
\item To the best of our knowledge, we are the first to identify the ``block effect" as a critical adversarial factor in deepfake detection, offering a new perspective on open-world deepfake detection.
\item To handle compressed deepfakes with limited paired data, we propose PLADA, comprising the B2E and ODA modules, addressing challenges from both data and model perspectives.
\item Extensive experiments validate PLADA's effectiveness across 26 datasets, demonstrating superior performance on OSNs compared to SoTA.
\end{itemize}

The preliminary version of this work was published in \citep{tao2025oddn} as an oral presentation. Compared to the conference paper, the current work introduces three key contributions. \emph{First}, we innovatively propose the concept of the ``block effect", a visual characteristic resulting from compression that has been overlooked by recent research. This observation provides new insights in the field. \emph{Second}, we present PLADA, a powerful new method that abandons the ``gradient correction" module used in previous methods. Instead, we propose a more effective and efficient mechanism that systematically incorporates individual guidance stages. This new approach excels in utilizing adversarial signals and addressing gradient conflicts, showing significant superiority over previous methods, especially in resource-constrained scenarios. \emph{Third}, we include more comprehensive studies and analyses of the newly proposed PLADA, enhancing the clarity and credibility of its performance and efficiency.


The sections are organized as follows. Section \ref{Sec:related-work} provides a comprehensive review of related works, including OSN preprocessing, deepfake generation, and detection. Section \ref{Sec:method} presents a detailed introduction to the proposed PLADA method. Section \ref{Sec:experiments} showcases the experimental results and analysis. Section \ref{Sec:conclusion} concludes the paper.

\section{Related Work}\label{Sec:related-work}
Deepfakes are typically categorized into two types: image manipulation and image generation. This study mainly focuses image generation, as it inherently encompasses aspects of the another. Here we briefly introduce information about influence of online social networks image process, recent advanced deepfakes techniques and their corresponding defensive approaches.

\subsection{OSN Preprocessing}
Online social networks (OSNs), including platforms such as Facebook, WeChat, and Weibo, significantly accelerates the spread and sharing of deepfake content. Numerous studies have explicitly pointed out that OSNs generally apply lossy compression techniques, specifically JPEG, to process user-uploaded images. A key concern is the block effect, a prominent artifact of JPEG compression in the spatial domain, which can easily be mistaken for forgery cues. As illustrated in Figure \ref{block_effect}, this effect, caused by dividing an image into $8\times8$ blocks during JPEG compression, introduces artifacts that often resemble deepfake traces, inevitably creating noisy for deepfake detection and analysis. Researchers such as Sun \citep{sun2016processing}, using Facebook as a case study, have conducted in-depth investigations and revealed that OSNs generally employ similar image processing pipelines. The typical image processing pipeline consists of four stages: (1) format conversion, (2) resizing, (3) image enhancement, and (4) JPEG compression. First, the pixel values of the uploaded image are cropped or normalized to comply with the standard pixel value range of [0, 255]. Next, if the image resolution exceeds a predefined threshold (e.g., 2048 pixels), it is resized. Subsequently, enhancement filters tailored to the image content are applied to optimize specific regions of the image. Finally, JPEG compression is performed using different quality factors, depending on transmission speed requirements. It is noteworthy that, while the preprocessing pipelines may vary slightly among different OSNs, JPEG remains a universally adopted technique. This highlights the urgent need to develop robust technologies capable of resisting and mitigating the challenges posed by JPEG compression.

\subsection{Deepfake Generation}
GANs use a training strategy rooted in a game between a generator and a discriminator \citep{GAN}, wherein the generator creates samples to deceive the discriminator, which aims to differentiate between real and fake samples. Over the years, several GAN variants have been proposed to generate realistic images, including BigGAN \citep{brock2018large} for class-conditional image generation and StyleGAN \citep{karras2021alias} for unconditional image generation. More recently, StyleGAN-T \citep{sauer2023stylegan} combined with Contrastive Language–Image Pre-training (CLIP) \citep{radford2021learning} for text-driven image generation. Despite GANs' ability to generate high-quality images with diverse attributes \citep{shen2020interpreting}, their manipulation is limited by the lack of an inference function. In contrast, DMs iteratively denoise random noise to generate images, with significant advantages in training stability and reduced mode collapse compared to GANs \citep{DM, dhariwal2021diffusion, balaji2022ediff}. DMs, including models like CLIP-integrated versions \citep{nichol2021glide, ramesh2022hierarchical} and Latent Diffusion Models (LDMs) \citep{rombach2022high}, have shown exceptional performance in image generation, with Stable Diffusion trained on a large-scale dataset and DiTs \citep{peebles2023scalable} introducing Transformer-based architectures for enhancement.

\subsection{Deepfake Detection}
To combat the misuse of deepfake technology, numerous studies \citep{li2024sa, duan2025test, guo2025language} have proposed various advanced strategies. Previous study \citep{wang2020cnn} has demonstrated that deepfake detection can achieve high effectiveness by leveraging a standard image classifier with simple fine-grained preprocessing. Similarly, other research \citep{ju2022fusing} has focused on integrating global and local image information to capture subtle forgery traces. Additionally, certain works \citep{frank2020leveraging, dzanic2020fourier} analyzed deepfakes in frequency domain, revealing that upsampling introduces identifiable frequency-domain artifacts, which inspired follow-up studies \citep{corvi2023detection, liu2022detecting} that utilized these artifacts for enhanced detection. Others \citep{xu2024learning, sun2025continual} combine spatial and temporal information for more accurate face forgery detection, while DIRE \citep{wang2023dire} introduces a novel image representation based on feature distance to improve deepfake detection.

\begin{figure*}[!t]
    \centering
    \includegraphics[width=\textwidth]{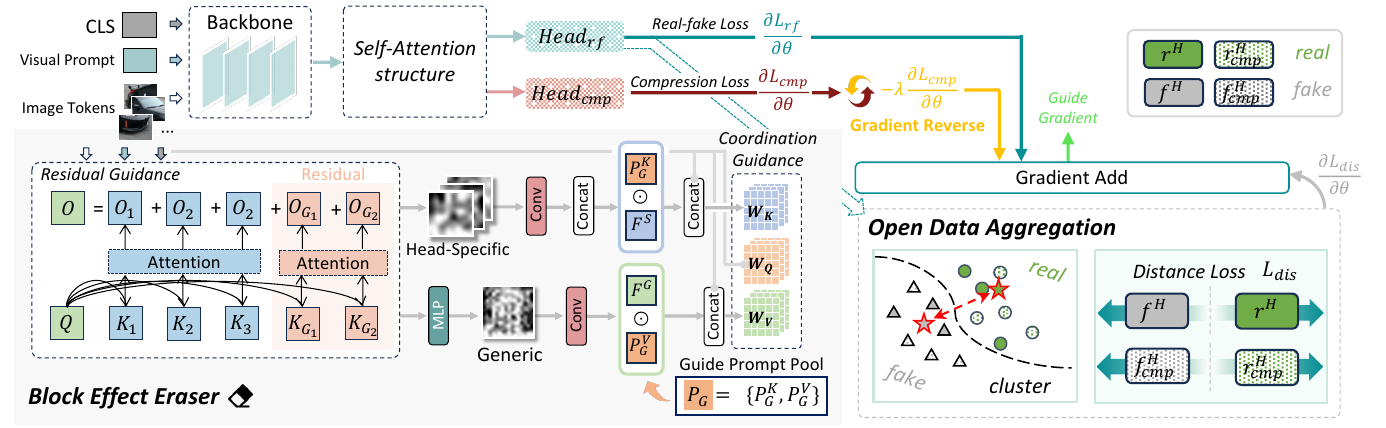} %
    \caption{\textbf{Overview of the PLADA Framework.} PLADA includes two core modules: B2E and ODA. Using gradient reversal, a guide gradient is generated by an auxiliary task and propagated alongside the gradient. B2E captures the guide gradient and stores it in guide prompts, redirecting attention to deepfake artifacts through ``attention shifting", distinct from conventional mechanisms. ODA leverages distributional correlations in unpaired and paired data to enhance performance.}
    \label{framework}
\end{figure*}

\section{Method}\label{Sec:method}
In this section, we delve into the details of the PLADA framework. We begin by defining and formalizing the problem, following the methodology of prior research \citep{tao2025oddn}. Next, we provide a comprehensive overview of the PLADA framework, mainly focusing on its structure and functionality. Finally, we offer specific details about the two core modules that form the foundation of the framework: B2E and ODA. 

\subsection{Problem Definition}
Given the scarcity of paired datasets containing both original-resolution deepfake images and their compressed variants, our approach takes a different direction from previous studies.


Consider a dataset $\mathcal{D} = \{(x_i, y_i)\}_{i=1}^{N}$, which encompasses two distinct categories of images: real images $x^r$ and deepfake images $x^f$. Each image is assigned a label $y \in \{0, 1\}$, where 0 indicates a real image, and 1 indicates a deepfake. To simulate a realistic setting, we randomly select a portion of the dataset $\mathcal{D}$ for JPEG compression, denoted as $\mathcal{P}_c$, with the compression level set to approximately 50\% image quality. This portion is limited to no more than 50\% of the dataset due to the scarcity of paired data. The uncompressed versions of these images are represented as $\mathcal{P}$, while the rest of the dataset is denoted as $\tilde{\mathcal{P}}$. Furthermore, we introduce an additional label $y^c \in \{0, 1\}$, where 0 denotes a uncompressed image, and 1 denotes a compressed one, to distinguish compressed images. As a result, the dataset is reformulated as $\mathcal{D}_{\text{train}} = \{(x_i, y_i, y^c_i)\}_{i=1}^{N}$, comprising the union of $\mathcal{P}$, $\mathcal{P}_c$, and $\tilde{\mathcal{P}}$. This formulation allows a clearer explanation of our methodology and ensures that our research setup aligns closely with real-world data scenarios.

At the inference stage, two distinct paradigms are considered: quality-aware and quality-agnostic inference. We define heterogeneous compression types as the use of different compression methods or parameters. In quality-aware inference, the test set images, denoted as $\mathcal{D}_{test}$, are compressed using same method and parameters as $\mathcal{P}_c$. Conversely, in quality-agnostic inference, the images in $\mathcal{D}_{test}$ are compressed using a variety of methods and parameters. This approach simulates an open-world scenario where the specific compression type is unknown. Our analysis reveals that the scarcity of paired data is a significant obstacle to enhancing model robustness, posing considerable challenges for a wide range of methods.

\subsection{Framework Overview}
As shown in Figure \ref{framework}, the novel PLADA framework consists of two core modules: B2E and ODA. The B2E module, incorporating two distinct knowledge injection mechanisms, redirects the backbone’s attention by suppressing block effect features and focusing exclusively on deepfake-related artifacts. By leveraging adversarial learning to activate and guide the B2E module, the framework masks and eliminates compression-related characteristics, allowing the detector to focus on deepfake-related information. In the ODA module, data manipulation is performed on two types of datasets: unpaired data and paired data. These datasets are aligned through unsupervised aggregation to ensure consistency.

\subsection{Block Effect Eraser (B2E) Module}
Although compression is widely recognized as a significant challenge in detecting deepfakes on OSNs, the primary factor, block effects, has been largely overlooked in recent researches. As demonstrated in preceding section, block effects and deepfake-related artifacts often share visual similarities, which severely hinder the model’s ability to accurately identify deepfake features. To tackle it, we propose a module called B2E, which is specifically designed to mitigate block effects by redirecting attention and improving the model’s inherent resilience to compression-induced artifacts. We refer to this deliberate and mandatory attention reconfiguration during the training phase as ``attention shifting''. Specifically, B2E module incorporates two complementary knowledge injection mechanisms: RG and CG, which synergistically collaborate to guide and facilitate the attention-shifting process within the backbone. In the subsequent sections, we provide a detailed exposition of these knowledge injection mechanisms, along with a thorough analysis of their functionality and empirical impact.

\textbf{Residual Guidance}. 
Typically, RG is similar to traditional MSA augmented with prompts, but it exhibits significant differences in computation formula. To clarify this, we first define the conventional MSA within the context of the PLADA framework, then explain the distinction.

Given a single input $\mathbf{x} \in \mathbb{R}^{(1 + 2+ L)\times D}$, it comprises a class token $P_{CLS} \in \mathbb{R}^D$, two visual tokens $P_{V} \in \mathbb{R}^{2\times D}$, and a sequence of image tokens $ I \in \mathbb{R}^{L\times D}$, where $D$ represents the embedding dimension and $L$ denotes the number of image patch tokens. These two visual tokens serve as learnable prompts, embedding deepfake information to facilitate artifact analysis. Within the backbone architecture, each intermediate layer integrates MSA and possesses an dedicated guide prompt pool denoted as $\mathcal{P}_G = \{(P_{G,i}^K, P_{G,i}^V)\}_{i=1}^{N}$, where $N$ represents the size of the pool. By employing linear projection functions, the input $x$ is transformed into query $\mathbf{h}_Q$, key $\mathbf{h}_K$, and value $\mathbf{h}_V$. Subsequently, $P_G^K$ and $P_G^V$ are randomly sampled from $\mathcal{P}_G$ and concatenated to $\mathbf{h}_K$ and $\mathbf{h}_V$, yielding $\widehat{\mathbf{h}_K}$ and $\widehat{\mathbf{h}_V}$. These modified key and value matrices, along with original query $\mathbf{h}_Q$, are then reshaped and fed into MSA, obtaining result $\mathbf{h}^{MSA}$. The aforementioned process is conventional MSA and can be formulated as follows:
\begin{equation}
\begin{aligned}
    \mathbf{h}^{MSA} &= MSA\left(f_Q(\mathbf{x}), \right. \\
    &\quad \left. [f_K(\mathbf{x}); P_G^K], \right. \\
    &\quad \left. [f_V(\mathbf{x}); P_G^V]\right)
\end{aligned} 
\end{equation}
where $MSA$ is the multi-head self-attention mechanism, $f_Q$, $f_K$, and $f_V$ denote the distinct linear functions for transforming $\mathbf{x}$ to query, key, and value respectively.

Conversely, to reduce excessive perturbation of the attention derived from the original image patch sequence while effectively supplementing additional guidance, RG adopts a novel residual knowledge injection mechanism. Specifically, RG first computes MSA result $\hat{\mathbf{h}}$ for the input $\mathbf{x}$. Then, $\mathbf{h}_Q$, along with $P_G^K$ and $P_G^V$, is fed into another MSA process, yielding $\overline{\mathbf{h}}$. Finally, $\hat{\mathbf{h}}$ and $\overline{\mathbf{h}}$ are simply summed to produce RG's output $\mathbf{h}^{RG}$. The above steps can be formulated as below.
\begin{equation}
    \hat{\mathbf{h}} = MSA(f_Q(\mathbf{x}),f_K(\mathbf{x}),f_V(\mathbf{x}))
\end{equation}
\begin{equation}
    \overline{\mathbf{h}} = MSA(f_Q(\mathbf{x}),f_K(P_G^K),f_V(P_G^V))
\end{equation}
\begin{equation}
    \mathbf{h}^{RG} = \hat{\mathbf{h}} + \overline{\mathbf{h}}
\end{equation}
This unique variant of MSA enables the knowledge embedded within the guide prompt to effectively lead the backbone to automatically concentrate on discovering deepfake-related artifacts. Simultaneously, it effectively eliminates misleading block effect traits in a coarse-grained manner.

\textbf{Coordination Guidance}. 
However, RG, derived from prefix tuning, inevitably inherits limitations in efficiently redirecting the pretrained backbone’s attention. Furthermore, several works have prove that it is crucial for the model to possess robust perception capabilities from both global and local perspectives. Hence, we introduce CG, another variant of MSA. Specifically, CG obtains generic features $F^G$ and head-specific features $F^S$ through the output of RG, followed by a multi-layer perceptron (MLP) and a series of convolution and concatenation operations. Subsequently, through a point-wise product to augment the feature vectors, the guide prompts $P_G^V$ and $P_G^K$  inject knowledge into  $F^G$ and $F^S$, producing $F^K$ and $F^V$ respectively. This knowledge helps distinguish between compression artifacts and accurate focus points. Then, $\mathbf{h}_K$ and $F^K$ are concatenated to form a new key $\hat{\mathbf{h}_K}$, and similarly, $\mathbf{h}_V$ and $F^V$ are concatenated to form a new value $\hat{\mathbf{h}_V}$. Finally, with $\mathbf{h}_Q$ as the query, the new key and value are fed into MSA. The aforementioned process of Coordination Guidance is outlined as follows:
\begin{equation}
\begin{aligned}
    F_K = P_G^K \odot &[g_1^S(\mathbf{h}_1^{RG}); \\
    & \left. g_2^S(\mathbf{h}_2^{RG}); \right. \\
    & \left. \ \ \ \ ...\ \ \ \ \ ; \right. \\
    & \left. g_H^S(\mathbf{h}_H^{RG})]\right.)
\end{aligned} 
\end{equation}
\begin{equation}
    F_V = P_G^V \odot g^G(MLP(\mathbf{h}^{RG}))
\end{equation}
\begin{equation}
    \mathbf{h}^{out} = MSA(\mathbf{h}_Q,[\mathbf{h}_K;F_K],[\mathbf{h}_V;F_V])
\end{equation}
Where $H$ denotes the number of heads in MSA, $MLP$ denotes the MLP layer, $\mathcal{G}^S = \{ g_1^S,g_2^S,...,g_H^S\}$ and $g^G$ represents convolution layers before forming $F^S$ and $F^G$ respectively, $\odot$ is point-wise product and $\mathbf{h}^{out}$ is our CG's output.

Under guidance, deepfake-related artifacts become more prominent in both the general features and head-specific features. As the key and value in MSA incorporate the enhanced head-specific and general features, MSA further facilitates a mapping of local attention to the global context in a bottom-up manner, effectively integrating multi-scale representations. Due to these intricate mechanisms, CG supports RG by enabling faster guidance storage and a more comprehensive attention-shifting process.

\textbf{Spark Guidance}. 
This raises several new questions: how can appropriate initial guidance information be constituted, and how can such information be embedded into the guide prompts? We hypothesize that compressed images inherently exhibit unique fingerprints, which are largely independent of the specific compression methods used. However, when a model is trained on datasets containing compressed images, it may inadvertently learn these fingerprint features, leading to bias and performance distortion. Our objective is to convert this fingerprint into a useful signal that can be stored in guide prompts, thereby redirecting the backbone network to shift its attention away from this compression fingerprint. To accomplish this, we leverage multi-task learning to extract the necessary guidance information. The loss functions for the two downstream tasks are defined as follows:
\begin{equation}
    \mathcal{L}_{cmp} = \mathcal{L}_{bce}(H_{cmp}(\mathbf{h}^{cmp}_{i}),\ y_i^c)
\end{equation}
\begin{equation}
    \mathcal{L}_{rf} = \mathcal{L}_{bce}(H_{rf}(\mathbf{h}^{rf}_{i}),\ y_i)
\end{equation} 
where \(\mathcal{L}_{bce}\) represents the binary cross-entropy loss, $H_{cmp}$ and $H_{rf}$ are the head of compression-discard branch and real/fake branch respectively, \(h^{cmp}_{i}\) and $h^{rf}_{i}$ denote the feature obtained from the self-attention block of different branches.

$\mathcal{L}_{tf}$ represents the primary task, aiming to aid the model in detecting deepfakes. $\mathcal{L}_{cmp}$ serves as the auxiliary task, responsible for detecting whether an image has been compressed. In detail, to mitigate data imbalance, after passing through the backbone, only the features of paired data are fed into the compression-discard branch, while the full dataset traverses real/fake branch. The gradients generated by $\mathcal{L}_{cmp}$ optimize the backbone's ability to identify compressed images. However, since we aim for the model to be insensitive to compression, a gradient reversal layer is applied to invert these gradients as they pass through. Consequently, when propagating through the network, the gradients in the backbone and the compression-discard branch are equal in magnitude but opposite in direction. This adversarial learning mechanism transforms the originally adverse signal into beneficial guidance information. Therefore, RG and CG can embed this information into guide prompts, utilizing it to help the backbone shift its attention and ultimately enhance its ability to distinguish deepfakes. 

RG and CG work together to constitute the Block Effect Eraser (B2E). However, we only use B2E to replace the traditional MSA in the shallow layers of the backbone, while the deeper layers rely solely on RG. By combining adversarial learning and multi-task learning, B2E extracts useful information from mixed gradients and embeds it into guide prompts. Consequently, this helps the backbone quickly shift its focus from deceptive compression features to truly deepfake-related artifacts. As a result, the backbone achieves a clearer focus and a stronger encoding capability.

\subsection{Open Data Aggregation (ODA) Module}
Previous studies \citep{le2023quality, tao2025oddn} used the Hilbert-Schmidt Independence Criterion (HSIC) to help models identify distributional correlations in paired data. But our experiments show that HSIC requires strict preconditions to work effectively. Specifically, it performs well with consistent compression types but negatively impacts detection and reduces performance when compression types vary. In contrast, our proposed ODA module offers a robust solution, capable of extracting valuable information from both paired and unpaired data, and have consistent performance across compression types. 

We firstly define state space $\Omega = \{\varphi, \phi\} \times \{\kappa, \overline{\kappa}\}$, where $\varphi/\phi$ represents real and fake classes, $\kappa/\overline{\kappa}$ denotes compressed and raw states respectively. Then, for a given batch of inputs $\mathbf{B} \in \mathbb{R}^{N\times D}$, ODA computes four aggregation centers: the real images denoted as $\mathbf{c}^{t}$, their compressed counterparts $\mathbf{c}^{t}_{cmp}$, the fake images $\mathbf{c}^{f}$, and the compressed fake images $\mathbf{c}^{f}_{cmp}$. 
The mathematical definitions of these four centers can be uniformly represented as follows:
\begin{equation}
\begin{aligned}
& C_{(\omega_1, \omega_2)} = \frac{\mathop{\Sigma}\limits_{\mathbf{x}_i \in \mathbf{B}}\Psi(\mathbf{h}_i^{H}) \odot \mathrm{I}[\mathbf{x}_i \in \mathbf{B}_{\omega_1}^{\omega_2}]}{N_{(\omega_1, \omega_2)}} \\
& \text{s.t.} \quad \mathop{\Sigma}\limits_{\omega_1, \omega_2} N_{(\omega_1, \omega_2)} = |\mathbf{B}|  \\
\end{aligned} 
\end{equation}
Where $\omega_1 \in \{\varphi, \phi\}$ is a class indicator, $\omega_2 \in \{ \kappa, \overline{\kappa}\}$ is a compression indicator, $\Psi$ is a function to intensify feature, $\mathbf{B}_{\omega_1}^{\omega_2}$ is a subset of $\mathbf{B}$ and belongs to specific state, $\mathrm{I}[\cdot]$ represents indicator function, $N_{(\omega_1, \omega_2)} = |B_{\omega_1}^{\omega_2}|$ is subset size, and $\mathbf{h}_i^{H}$ denotes features extracted from the attention branch.

Subsequently, we augment the separation among these clustering centers to enhance the ability of distinguishing between real and fake images, enabling the detector to identify them more accurately. Specifically, to ensure our model effectively classify deepfakes, even in the compressed forms, we amplify the distance $s$ between the cluster centers \(\mathbf{c}^{t}\) and \(\mathbf{c}^{f}\), and likewise, the distance \(s_{cmp}\) between the compressed cluster centers \(\mathbf{c}^{t}_{cmp}\) and \(\mathbf{c}^{f}_{cmp}\). The formulas are uniformly provided as below:
\begin{equation}
    \begin{split}
    &  S(\omega_{1}^i,\omega_{1}^j,\hat\kappa) = \frac{1}{1 + \tau \cdot D_{p} \left( C_{(\omega_{1}^i,\hat{\kappa})}, C_{(\omega_{1}^j,\hat{\kappa})} \right)}\\
    &\text{s.t.}\quad  \left\{\begin{array}{lc}
    D_{p} \left(\mathbf{a},\mathbf{b}\right) = \left(\overset{d}{\mathop{\Sigma}\limits_{i = 1}}\lambda_i\left|\frac{a_i - b_i}{\sigma_i + \epsilon} \right|^p \right)^{1/p}\\
    \lambda_i = \mathrm{exp}\left(-\frac{(a_i-\mu_i)^2}{2\sigma_i^2} \right)\\
    \end{array}\right.
    \end{split}
\end{equation}
where $d$ denotes the dimension of the aggregation centers, $\tau > 0$ is a temperature adjustment coefficient, $p\in[1,\infty)$ represents variable order, $D_{p}$ is a distance functional.
\begin{equation}
\begin{aligned}
  \mathcal{L}_{dis} &= \mathop{\Sigma}\limits_{\omega_2 \in \{ \kappa, \overline{\kappa}\}}S (\varphi,\phi,\omega_{2})\\ & + 
\beta \mathop{\Sigma}\limits_{\omega_1 \in \{\varphi, \phi\}} \mathrm{H} \left(C _{(\omega_{1},\kappa)}, C_{(\omega_{1},\bar{\kappa})} \right)  \\
\text{s.t.} \quad \beta &= \sigma\left(\gamma\cdot\frac{\partial \mathcal{L}_{dis}}{\partial\Theta}\right) \in (0,1) \\
\end{aligned} 
\end{equation}
Where $\sigma(\cdot)$ represents sigmoid function, $\gamma$ is a predefined coefficient, $\mathrm{H}(\cdot,\cdot)$ denotes Hilbert-Schmidt Independence Criterion divergence.

Consequently, the summation of $S_{cmp}$ and $S$ forms an alignment loss, denoted as $\mathcal{L}_{dis}$. This loss function serves a dual purpose: it increases the separation between real and fake images while promoting the cohesion of images within the same class. Unlike HSIC, ODA is not limited to exclusive interactions with paired data. During the gradient updates, $S_{cmp}$ and $S$ interact and mutually constrain each other, enabling the model to learn refined correlations among $P$, $P_c$, and $\tilde{P}$. Hence, this enables model to possess robust encoding capabilities. In essence, ODA places additional emphasis on unpaired data, an aspect largely neglected by mainstream research, while preserving and even enhancing its ability in uncovering correlations within paired data. 

Overall, the final loss function for the training process is a weighted sum of the above losses:
\begin{equation}
    \mathcal{L}_{all} = \mathcal{L}_{tf} + \mathcal{L}_{cmp} + \alpha\mathcal{L}_{dis} + \mathcal{R}(\Theta)
\end{equation}
Where $\alpha$ is predefined coefficient, $\mathcal{R}(\Theta)$  represents structural loss of model.

\begin{table*}[!t]
\scriptsize
\centering
\caption{\textbf{Quality-Aware Experimental Results} on the \textit{ForenSynths} and \textit{GANGen-Detection} Datasets under a 4-Class Training Data Setting. The symbol \dag\ indicates results from the original paper or previous work, while \ddag\ denotes results obtained by retraining the model with pretrained weights. \textbf{Bold} highlights the best performance, and \underline{underlining} indicates the second-best performance.}
\label{4class_17GANs_Quality_aware}

\setlength{\tabcolsep}{0.48mm}{
\begin{tabular}{l| c c c c c c c c c| c c c c c c c c |c}
\toprule
\multirow{3}{*}{\footnotesize Methods} 
& \multicolumn{9}{c|}{\footnotesize \textit{GANGen-Detection} Dataset \citep{chuangchuangtan-GANGen-Detection}} 
& \multicolumn{8}{c|}{\footnotesize \textit{ForenSynths} Dataset \citep{wang2020cnn}} 
& \multirow{3}{*}{\makecell[c]{\footnotesize Mean \\ \footnotesize Acc}} \\

\cmidrule(lr){2-10} \cmidrule(lr){11-18}
& \makecell[c]{\footnotesize Info-\\\footnotesize GAN} 
& \makecell[c]{\footnotesize BE-\\\footnotesize GAN} 
& \makecell[c]{\footnotesize Cram-\\\footnotesize GAN} 
& \makecell[c]{\footnotesize Att-\\\footnotesize GAN} 
& \makecell[c]{\footnotesize MMD-\\\footnotesize GAN} 
& \makecell[c]{\footnotesize Rel-\\\footnotesize GAN} 
& \makecell[c]{\footnotesize S3-\\\footnotesize GAN} 
& \makecell[c]{\footnotesize SNG-\\\footnotesize GAN} 
& \makecell[c]{\footnotesize STG-\\\footnotesize GAN} 
& \makecell[c]{\footnotesize Pro-\\\footnotesize GAN} 
& \makecell[c]{\footnotesize Style-\\\footnotesize GAN} 
& \makecell[c]{\footnotesize Style-\\\footnotesize GAN2} 
& \makecell[c]{\footnotesize Big-\\\footnotesize GAN} 
& \makecell[c]{\footnotesize Cycle-\\\footnotesize GAN} 
& \makecell[c]{\footnotesize Star-\\\footnotesize GAN} 
& \makecell[c]{\footnotesize Gau-\\\footnotesize GAN} 
& \makecell[c]{\footnotesize Deep-\\\footnotesize fake} 
& \\

\midrule
\footnotesize \href{https://openaccess.thecvf.com/content_ICCV_2019/papers/Rossler_FaceForensics_Learning_to_Detect_Manipulated_Facial_Images_ICCV_2019_paper.pdf}{FF++\dag (2019)} & \footnotesize69.5 & \footnotesize26.9 & \footnotesize80.3 & \footnotesize66.8 & \footnotesize79.2 & \footnotesize69.9 & \footnotesize56.2 & \footnotesize75.1 & \footnotesize84.4 & \footnotesize93.6 & \footnotesize62.5 & \footnotesize60.8 & \footnotesize58.5 & \footnotesize80.9 & \footnotesize78.5 & \footnotesize71.0 & \footnotesize52.8 & \footnotesize68.7\\

\footnotesize \href{https://arxiv.org/pdf/2007.09355}{F3Net\dag (2020)} & \footnotesize61.0 & \footnotesize41.9 & \footnotesize65.8 & \footnotesize52.9 & \footnotesize63.8 & \footnotesize55.5 & \footnotesize53.8 & \footnotesize59.6 & \footnotesize71.5 & \footnotesize92.2 & \footnotesize76.0 & \footnotesize59.1 & \footnotesize55.9 & \footnotesize57.9 & \footnotesize71.8 & \footnotesize66.0 & \footnotesize52.1 & \footnotesize62.4\\

\footnotesize \href{https://openaccess.thecvf.com/content/CVPR2021/papers/Zhao_Multi-Attentional_Deepfake_Detection_CVPR_2021_paper.pdf}{MAT\dag (2021)} & \footnotesize57.9 & \footnotesize46.9 & \footnotesize64.2 & \footnotesize50.8 & \footnotesize63.4 & \footnotesize52.4 & \footnotesize52.1 & \footnotesize56.2 & \footnotesize61.8 & \footnotesize90.8 & \footnotesize54.2 & \footnotesize53.9 & \footnotesize52.4 & \footnotesize73.1 & \footnotesize61.4 & \footnotesize64.8 & \footnotesize51.2 & \footnotesize59.5\\

\footnotesize \href{https://openaccess.thecvf.com/content/CVPR2022/papers/Shiohara_Detecting_Deepfakes_With_Self-Blended_Images_CVPR_2022_paper.pdf}{SBI\dag (2022)} & \footnotesize60.2 & \footnotesize55.7 & \footnotesize74.4 & \footnotesize50.2 & \footnotesize67.1 & \footnotesize54.6 & \footnotesize61.4 & \footnotesize53.0 & \footnotesize57.2 & \footnotesize96.0 & \footnotesize57.4 & \footnotesize53.0 & \footnotesize55.4 & \footnotesize77.6 & \footnotesize60.1 & \footnotesize74.9 & \footnotesize50.6 & \footnotesize62.5\\

\footnotesize \href{https://openaccess.thecvf.com/content/ICCV2023/papers/Le_Quality-Agnostic_Deepfake_Detection_with_Intra-model_Collaborative_Learning_ICCV_2023_paper.pdf}{QAD\dag (2023)} & \footnotesize79.9 & \footnotesize37.5 & \footnotesize79.5 & \footnotesize67.4 & \footnotesize76.8 & \footnotesize71.7 & \footnotesize58.0 & \footnotesize79.0 & \footnotesize83.5 & \footnotesize92.7 & \footnotesize64.7 & \footnotesize68.7 & \footnotesize64.0 & \footnotesize81.8 & \footnotesize80.3 & \footnotesize66.3 & \footnotesize52.9 & \footnotesize70.9\\

\footnotesize \href{https://openaccess.thecvf.com/content/CVPR2023/papers/Ojha_Towards_Universal_Fake_Image_Detectors_That_Generalize_Across_Generative_Models_CVPR_2023_paper.pdf}{Ojha\ddag (2023)} & \footnotesize70.9 & \footnotesize57.6 & \footnotesize68.5 & \footnotesize59.1 & \footnotesize68.9 & \footnotesize63.9 & \footnotesize71.9 & \footnotesize64.2 & \footnotesize60.0 & \footnotesize89.8 & \footnotesize66.4 & \footnotesize63.7 & \footnotesize67.7 & \footnotesize84.2 & \footnotesize76.3 & \footnotesize87.0 & \footnotesize63.1 & \footnotesize69.9\\

\footnotesize \href{https://openaccess.thecvf.com/content/CVPR2024/papers/Tan_Rethinking_the_Up-Sampling_Operations_in_CNN-based_Generative_Network_for_Generalizable_CVPR_2024_paper.pdf}{NPR\ddag (2024)} & \footnotesize63.4 & \footnotesize45.9 & \footnotesize54.2 & \footnotesize62.1 & \footnotesize55.5 & \footnotesize70.1 & \footnotesize55.7 & \footnotesize55.8 & \footnotesize59.6 & \footnotesize80.5 & \footnotesize67.4 & \footnotesize64.2 & \footnotesize62.6 & \footnotesize70.7 & \footnotesize67.5 & \footnotesize58.0 & \footnotesize53.6 & \footnotesize61.6\\

\footnotesize \href{https://openaccess.thecvf.com/content/CVPR2024/papers/Liu_Forgery-aware_Adaptive_Transformer_for_Generalizable_Synthetic_Image_Detection_CVPR_2024_paper.pdf}{FaT\ddag (2024)} & \footnotesize50.9 & \footnotesize57.1 & \footnotesize48.9 & \footnotesize46.6 & \footnotesize48.9 & \footnotesize46.7 & \footnotesize50.2 & \footnotesize50.2 & \footnotesize50.6 & \footnotesize50.0 & \footnotesize50.3 & \footnotesize51.7 & \footnotesize52.2 & \footnotesize51.7 & \footnotesize50.2 & \footnotesize52.2 & \footnotesize52.4 & \footnotesize50.7\\

\midrule
\footnotesize \href{https://doi.org/10.1609/aaai.v39i1.32063}{ODDN\dag (2025)} & \footnotesize80.6 & \footnotesize38.6 & \footnotesize80.7 & \footnotesize65.8 & \footnotesize78.8 & \footnotesize71.1 & \footnotesize60.5 & \footnotesize76.7 & \footnotesize85.8 & \footnotesize94.0 & \footnotesize67.7 & \footnotesize69.9 & \footnotesize66.7 & \footnotesize84.9 & \footnotesize80.5 & \footnotesize75.2 & \footnotesize54.2 & \footnotesize\underline{72.6}\\

\rowcolor{gray!20}
\footnotesize PLADA(\textbf{ours}) & \footnotesize77.6 & \footnotesize73.8 & \footnotesize79.7 & \footnotesize66.2 & \footnotesize79.3 & \footnotesize67.6 & \footnotesize76.5 & \footnotesize75.3 & \footnotesize77.8 & \footnotesize92.8 & \footnotesize68.5 & \footnotesize64.7 & \footnotesize71.6 & \footnotesize83.5 & \footnotesize83.5 & \footnotesize90.5 & \footnotesize74.0 & \footnotesize\textbf{76.7}\\

\bottomrule
\end{tabular}}
\end{table*}

\begin{table*}[!t]
\scriptsize
\centering
\caption{\textbf{Quality-Aware Experimental Results} on the \textit{Ojha-test} Dataset \citep{ojha2023towards} under a 4-Class Training Data Setting. The marks are consistent with those in Table 1.}
\label{4class_8DMs_Quality_aware}

\setlength{\tabcolsep}{0.58mm}{
\begin{tabular}{l | c c | c c | c c | c c | c c | c c | c c | c c | c c}
\toprule
 \multirow{2}{*}{\footnotesize Methods} 
& \multicolumn{2}{c|}{\footnotesize DALLE} 
& \multicolumn{2}{c|}{\footnotesize Glide\_100\_10}
& \multicolumn{2}{c|}{\footnotesize Glide\_100\_27}
& \multicolumn{2}{c|}{\footnotesize Glide\_50\_27}
& \multicolumn{2}{c|}{\footnotesize ADM}
& \multicolumn{2}{c|}{\footnotesize LDM\_100}
& \multicolumn{2}{c|}{\footnotesize LDM\_200}
& \multicolumn{2}{c|}{\footnotesize LDM\_200\_cfg}
& \multicolumn{2}{c}{\footnotesize Mean} \\


\cmidrule{2-19}
& \multicolumn{2}{c|}{\footnotesize Acc. \ \ \footnotesize A.P.}  
& \multicolumn{2}{c|}{\footnotesize Acc. \ \ \footnotesize A.P.}   
& \multicolumn{2}{c|}{\footnotesize Acc. \ \ \footnotesize A.P.}  
& \multicolumn{2}{c|}{\footnotesize Acc. \ \ \footnotesize A.P.} 
& \multicolumn{2}{c|}{\footnotesize Acc. \ \ \footnotesize A.P.}  
& \multicolumn{2}{c|}{\footnotesize Acc. \ \ \footnotesize A.P.}   
& \multicolumn{2}{c|}{\footnotesize Acc. \ \ \footnotesize A.P.}  
&\multicolumn{2}{c|}{\footnotesize Acc. \ \ \footnotesize A.P.} 
& \multicolumn{2}{c}{\footnotesize Acc. \ \ \footnotesize A.P.} \\

\midrule

\footnotesize \href{https://openaccess.thecvf.com/content_ICCV_2019/papers/Rossler_FaceForensics_Learning_to_Detect_Manipulated_Facial_Images_ICCV_2019_paper.pdf}{FF++\ddag (2019)} 
& \multicolumn{2}{c|}{\footnotesize 54.9 \ \ \footnotesize 68.0}  
& \multicolumn{2}{c|}{\footnotesize 52.7 \ \ \footnotesize 64.1}   
& \multicolumn{2}{c|}{\footnotesize 53.1 \ \ \footnotesize 63.0}  
& \multicolumn{2}{c|}{\footnotesize 53.0 \ \ \footnotesize 64.8} 
& \multicolumn{2}{c|}{\footnotesize 55.4 \ \ \footnotesize 72.7}  
& \multicolumn{2}{c|}{\footnotesize 55.6 \ \ \footnotesize 71.8}   
& \multicolumn{2}{c|}{\footnotesize 55.8 \ \ \footnotesize 71.3}  
&\multicolumn{2}{c|}{\footnotesize 52.2 \ \ \footnotesize 62.2} 
& \multicolumn{2}{c}{\footnotesize 54.1 \ \ \footnotesize 67.3} \\

\footnotesize \href{https://arxiv.org/pdf/2007.09355}{F3Net\ddag (2020)} 
& \multicolumn{2}{c|}{\footnotesize 55.1 \ \ \footnotesize 61.4}  
& \multicolumn{2}{c|}{\footnotesize 53.1 \ \ \footnotesize 59.6}   
& \multicolumn{2}{c|}{\footnotesize 53.4 \ \ \footnotesize 58.7}  
& \multicolumn{2}{c|}{\footnotesize 54.6 \ \ \footnotesize 61.2} 
& \multicolumn{2}{c|}{\footnotesize 58.0 \ \ \footnotesize 71.1}  
& \multicolumn{2}{c|}{\footnotesize 54.1 \ \ \footnotesize 60.3}   
& \multicolumn{2}{c|}{\footnotesize 54.4 \ \ \footnotesize 59.8}  
&\multicolumn{2}{c|}{\footnotesize 51.4 \ \ \footnotesize 54.2} 
& \multicolumn{2}{c}{\footnotesize 54.2 \ \ \footnotesize 60.8} \\

\footnotesize \href{https://openaccess.thecvf.com/content/CVPR2021/papers/Zhao_Multi-Attentional_Deepfake_Detection_CVPR_2021_paper.pdf}{MAT\dag (2021)} 
& \multicolumn{2}{c|}{\footnotesize 64.4 \ \ \footnotesize 48.6}  
& \multicolumn{2}{c|}{\footnotesize 63.0 \ \ \footnotesize 47.3}   
& \multicolumn{2}{c|}{\footnotesize 63.1 \ \ \footnotesize 47.1}  
& \multicolumn{2}{c|}{\footnotesize 62.4 \ \ \footnotesize 47.2} 
& \multicolumn{2}{c|}{\footnotesize 49.4 \ \ \footnotesize 38.9}  
& \multicolumn{2}{c|}{\footnotesize 66.7 \ \ \footnotesize 47.7}   
& \multicolumn{2}{c|}{\footnotesize 66.5 \ \ \footnotesize 47.7}  
&\multicolumn{2}{c|}{\footnotesize 65.8 \ \ \footnotesize 48.2}
& \multicolumn{2}{c}{\footnotesize 62.6 \ \ \footnotesize 46.6} \\

\footnotesize \href{https://openaccess.thecvf.com/content/CVPR2022/papers/Shiohara_Detecting_Deepfakes_With_Self-Blended_Images_CVPR_2022_paper.pdf}{SBI\ddag (2022)} 
& \multicolumn{2}{c|}{\footnotesize 52.6 \ \ \footnotesize 57.4}  
& \multicolumn{2}{c|}{\footnotesize 53.3 \ \ \footnotesize 63.1}   
& \multicolumn{2}{c|}{\footnotesize 54.1 \ \ \footnotesize 62.9}  
& \multicolumn{2}{c|}{\footnotesize 54.4 \ \ \footnotesize 64.3} 
& \multicolumn{2}{c|}{\footnotesize 55.8 \ \ \footnotesize 65.5}  
& \multicolumn{2}{c|}{\footnotesize 58.1 \ \ \footnotesize 71.4}   
& \multicolumn{2}{c|}{\footnotesize 57.3 \ \ \footnotesize 70.5}  
&\multicolumn{2}{c|}{\footnotesize 51.7 \ \ \footnotesize 57.1} 
& \multicolumn{2}{c}{\footnotesize 54.6 \ \ \footnotesize 64.0} \\

\footnotesize \href{https://openaccess.thecvf.com/content/ICCV2023/papers/Le_Quality-Agnostic_Deepfake_Detection_with_Intra-model_Collaborative_Learning_ICCV_2023_paper.pdf}{QAD\dag (2023)} 
& \multicolumn{2}{c|}{\footnotesize 53.3 \ \ \footnotesize 54.5}  
& \multicolumn{2}{c|}{\footnotesize 60.4 \ \ \footnotesize 66.6}   
& \multicolumn{2}{c|}{\footnotesize 60.9 \ \ \footnotesize 66.9}  
& \multicolumn{2}{c|}{\footnotesize 61.7 \ \ \footnotesize 68.5} 
& \multicolumn{2}{c|}{\footnotesize 57.8 \ \ \footnotesize 58.6}  
& \multicolumn{2}{c|}{\footnotesize 60.6 \ \ \footnotesize 66.9}   
& \multicolumn{2}{c|}{\footnotesize 61.0 \ \ \footnotesize 67.8}  
&\multicolumn{2}{c|}{\footnotesize 50.9 \ \ \footnotesize 50.4} 
& \multicolumn{2}{c}{\footnotesize 58.3 \ \ \footnotesize 62.5} \\

\footnotesize \href{https://openaccess.thecvf.com/content/CVPR2023/papers/Ojha_Towards_Universal_Fake_Image_Detectors_That_Generalize_Across_Generative_Models_CVPR_2023_paper.pdf}{Ojha\ddag (2023)} 
& \multicolumn{2}{c|}{\footnotesize 62.3 \ \ \footnotesize 69.2}  
& \multicolumn{2}{c|}{\footnotesize 75.6 \ \ \footnotesize 84.2}   
& \multicolumn{2}{c|}{\footnotesize 74.5 \ \ \footnotesize 83.1}  
& \multicolumn{2}{c|}{\footnotesize 76.2 \ \ \footnotesize 84.7} 
& \multicolumn{2}{c|}{\footnotesize 75.1 \ \ \footnotesize 70.5}  
& \multicolumn{2}{c|}{\footnotesize 76.8 \ \ \footnotesize 84.2}   
& \multicolumn{2}{c|}{\footnotesize 78.1 \ \ \footnotesize 85.1}  
&\multicolumn{2}{c|}{\footnotesize 59.8 \ \ \footnotesize 64.2} 
& \multicolumn{2}{c}{\footnotesize \underline{71.1} \ \ \footnotesize \underline{78.1}} \\

\footnotesize \href{https://openaccess.thecvf.com/content/CVPR2024/papers/Tan_Rethinking_the_Up-Sampling_Operations_in_CNN-based_Generative_Network_for_Generalizable_CVPR_2024_paper.pdf}{NPR\ddag (2024)} 
& \multicolumn{2}{c|}{\footnotesize 65.5 \ \ \footnotesize 64.0}  
& \multicolumn{2}{c|}{\footnotesize 58.3 \ \ \footnotesize 54.9}   
& \multicolumn{2}{c|}{\footnotesize 58.4 \ \ \footnotesize 55.0}  
& \multicolumn{2}{c|}{\footnotesize 58.9 \ \ \footnotesize 55.9} 
& \multicolumn{2}{c|}{\footnotesize 81.3 \ \ \footnotesize 79.4}  
& \multicolumn{2}{c|}{\footnotesize 62.1 \ \ \footnotesize 57.1}   
& \multicolumn{2}{c|}{\footnotesize 61.9 \ \ \footnotesize 57.3}  
&\multicolumn{2}{c|}{\footnotesize 58.0 \ \ \footnotesize 55.0} 
& \multicolumn{2}{c}{\footnotesize 63.0 \ \ \footnotesize 59.8} \\

\footnotesize \href{https://openaccess.thecvf.com/content/CVPR2024/papers/Liu_Forgery-aware_Adaptive_Transformer_for_Generalizable_Synthetic_Image_Detection_CVPR_2024_paper.pdf}{FaT\ddag (2024)} 
& \multicolumn{2}{c|}{\footnotesize 52.6 \ \ \footnotesize 51.4}  
& \multicolumn{2}{c|}{\footnotesize 52.3 \ \ \footnotesize 51.2}   
& \multicolumn{2}{c|}{\footnotesize 51.7 \ \ \footnotesize 50.9}  
& \multicolumn{2}{c|}{\footnotesize 53.5 \ \ \footnotesize 52.0} 
& \multicolumn{2}{c|}{\footnotesize 57.9 \ \ \footnotesize 55.4}  
& \multicolumn{2}{c|}{\footnotesize 50.4 \ \ \footnotesize 50.2}   
& \multicolumn{2}{c|}{\footnotesize 49.8 \ \ \footnotesize 49.9}  
&\multicolumn{2}{c|}{\footnotesize 49.0 \ \ \footnotesize 49.5} 
& \multicolumn{2}{c}{\footnotesize 52.1 \ \ \footnotesize 51.3} \\

\midrule
\footnotesize \href{https://doi.org/10.1609/aaai.v39i1.32063}{ODDN\dag (2025)} 
& \multicolumn{2}{c|}{\footnotesize 54.4 \ \ \footnotesize 62.1}  
& \multicolumn{2}{c|}{\footnotesize 53.4 \ \ \footnotesize 60.2}   
& \multicolumn{2}{c|}{\footnotesize 52.6 \ \ \footnotesize 58.1}  
& \multicolumn{2}{c|}{\footnotesize 54.0 \ \ \footnotesize 61.7} 
& \multicolumn{2}{c|}{\footnotesize 54.9 \ \ \footnotesize 61.8}  
& \multicolumn{2}{c|}{\footnotesize 51.5 \ \ \footnotesize 55.3}   
& \multicolumn{2}{c|}{\footnotesize 51.4 \ \ \footnotesize 54.2}  
&\multicolumn{2}{c|}{\footnotesize 52.2 \ \ \footnotesize 53.4} 
& \multicolumn{2}{c}{\footnotesize 53.0 \ \ \footnotesize 58.4} \\

\rowcolor{gray!20}
\footnotesize PLADA(\textbf{ours}) 
& \multicolumn{2}{c|}{\footnotesize 67.1 \ \ \footnotesize 79.2}  
& \multicolumn{2}{c|}{\footnotesize 71.9 \ \ \footnotesize 85.4}   
& \multicolumn{2}{c|}{\footnotesize 73.9 \ \ \footnotesize 86.0}  
& \multicolumn{2}{c|}{\footnotesize 74.1 \ \ \footnotesize 86.1} 
& \multicolumn{2}{c|}{\footnotesize 70.8 \ \ \footnotesize 85.0}  
& \multicolumn{2}{c|}{\footnotesize 84.3 \ \ \footnotesize 93.7}   
& \multicolumn{2}{c|}{\footnotesize 84.4 \ \ \footnotesize 93.6}  
&\multicolumn{2}{c|}{\footnotesize 60.7 \ \ \footnotesize 72.5} 
& \multicolumn{2}{c}{\footnotesize \textbf{73.4} \ \ \footnotesize \textbf{85.2}} \\

\bottomrule
\end{tabular}}
\end{table*}

\section{Experiment}\label{Sec:experiments}
\subsection{Experimental Setup}
\textbf{Datasets and Metrics.} Consistent with standard training and testing practices \citep{wang2020cnn, tan2024rethinking}, we train our model using the training set from the \textit{ForenSynths} dataset, following established baselines \citep{wang2020cnn}. For GAN-based evaluation, we assembled a comprehensive collection of 17 widely used datasets. Specifically, the first eight subsets are sourced from the \textit{ForenSynths} dataset, covering images generated by eight diverse generative adversarial models. The next nine subsets are extracted from the \textit{GANGen-Detection} \citep{chuangchuangtan-GANGen-Detection} dataset, including images generated by an additional nine GANs. For DM-based evaluation, the \textit{Ojha-test} dataset is sourced from \citep{ojha2023towards}, while another DM-based dataset, \textit{DiffusionForensics}, is taken from \citep{tan2024rethinking}. We use the commonly adopted evaluation metrics of accuracy (Acc.) and average precision (A.P.).

\textbf{Baseline and Implementation.} Based on the challenges identified and the architecture of our backbone model,  we limit our baseline comparisons to three categories of methods: 
those designed to resist OSN manipulations, those claiming robustness to compression, and naive CLIP along with its variants tailored to defend against deepfake images. 
Consequently, we incorporate FF++ \citep{rossler2019faceforensics++}, F3Net \citep{qian2020thinking}, MAT \citep{zhao2021multi}, SBI \citep{shiohara2022detecting}, QAD \citep{le2023quality}, Ojha \citep{ojha2023towards}, NPR \citep{tan2024rethinking}, FaT \citep{liu2024FaT} and ODDN \citep{tao2025oddn} as competitors. Unless stated otherwise, all the above methods are retrained using our experimental settings.

We utilize CLIP as our backbone and substitute its first two MSA layers with B2E, while the remaining MSA layers solely employ RG. For selecting guide prompts, we randomly sample them during the training phase and compute the average of these prompts within the pool for the inference phase. Regarding $P_{cls}$, $P_V$, and $\mathcal{P}_G$, we randomly init them with gaussian distribution. We adopt Adam as our optimizer, configuring it with a learning rate of \(2 \times 10^{-4}\) and a batch size of 32. The hyper-parameter $\alpha$ is set to 0.004 following standard configuration \citep{tao2025oddn}. All experiments are conducted using PyTorch on two NVIDIA GeForce RTX 4090 GPUs.

\begin{figure*}[!t]
    \centering
    \includegraphics[scale=0.48]{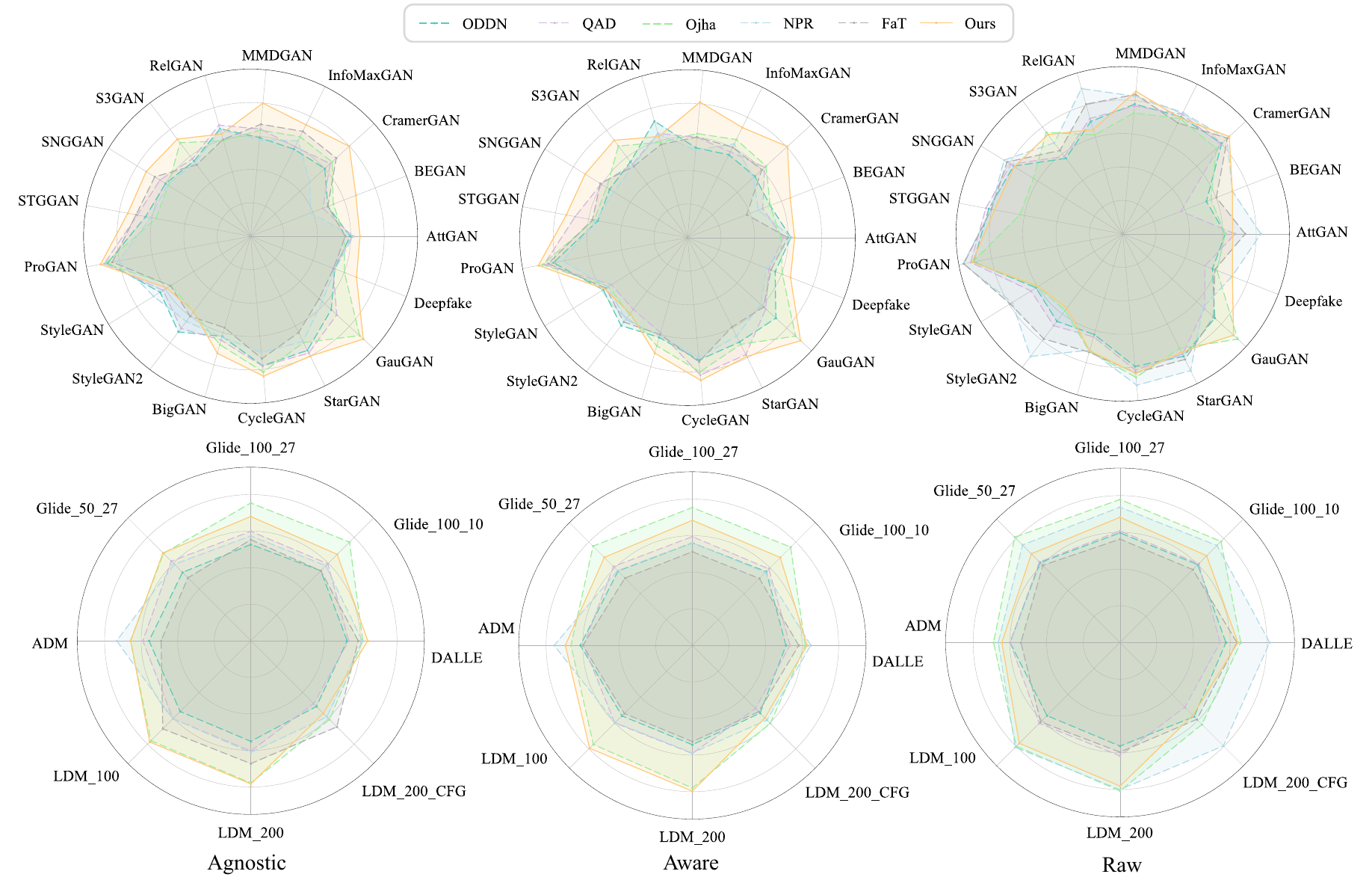} 
    \caption{The \textbf{Quality-Aware}, \textbf{Quality-Agnostic}, and \textbf{Raw Image} Experimental Results under a \textbf{2-Class Training Data Setting} on the \textit{DiffusionForensics}, \textit{ForenSynths}, and \textit{GANGen-Detection} Datasets.}
    \label{2class}
\end{figure*}

\subsection{Quality-aware Evaluation}
Following the established benchmark, we use a 4-Class subset consisting of ``car", ``cat", ``chair", and ``horse", as well as a 2-Class subset containing ``chair" and ``horse" from the \textit{ForenSynths} dataset. To simulate the data composition in simple OSN scenario, 20\% of the dataset is automatically compressed using common OSN operations with a constant quantization parameter of 50, creating paired data. These paired data, along with the remaining 80\% unpaired data, are utilized for model training. During the inference phase, test images compressed with the same quantization parameter are input into the model to analysis whether it has effectively utilized the prior knowledge. Notably, if baselines are specifically designed for paired data, unpaired data should also be incorporated into their training using an appropriate strategy, rather than being excluded. This ensures a fair and comprehensive comparison.

As shown in Table \ref{4class_17GANs_Quality_aware}, Table \ref{4class_8DMs_Quality_aware}, and Figure \ref{5DMs}, our proposed method, PLADA, exhibits superior performance on GAN datasets while maintaining a competitive edge on DM datasets. When evaluated across 17 GANs, PLADA attains a peak mean accuracy of 76.7\%, underscoring its effectiveness in this classification task. ODDA trails behind with an average accuracy of 72.6\%, revealing a notable performance gap of over 4\% compared to PLADA. Although PLADA incorporates more optimization parameters than Ojha, inevitably introducing domain adaptation challenges, it does not suffer from performance degradation in the context of DMs, even performing better. Thereby, it still preserve the inherent generalization of the CLIP backbone.

To showcase the generalization of PLADA in real-world scenarios, we conducted quality-aware experiments involving two classes. As depicted in Figure \ref{2class}, PLADA occupies the largest area and encapsulates other comparative approaches in many datasets, indicating its excellent generalization even with limited data. In Figure \ref{5DMs}, PLADA surpasses other baseline methods and closely trails the naive CLIP model. This suggests that PLADA does not experience a severe performance decline, maintaining its capability of adapting to other domains in such complex condition. Given that training data often cannot fully represent real-world conditions, above characteristics are crucial, as they ensure that PLADA can be effectively utilized in practical applications.

\begin{table*}[!t]
\scriptsize
\centering
\caption{The \textbf{Quality-Agnostic} Experimental Results under a \textbf{4-Class Training Data Setting} on the \textit{ForenSynths} and \textit{GANGen-Detection} Datasets. The marks are consistent with those in Table 1.}
\label{4class-17GANs-Quality-agnostic}

\setlength{\tabcolsep}{0.48mm}{
\begin{tabular}{l| c c c c c c c c c| c c c c c c c c |c}
\toprule
\multirow{3}{*}{\footnotesize Methods} 
& \multicolumn{9}{c|}{\footnotesize \textit{GANGen-Detection} \citep{chuangchuangtan-GANGen-Detection}} 
& \multicolumn{8}{c|}{\footnotesize \textit{ForenSynths} \citep{wang2020cnn}} 
& \multirow{3}{*}{\makecell[c]{\footnotesize Mean \\ \footnotesize Acc}} \\

\cmidrule(lr){2-10} \cmidrule(lr){11-18}
& \makecell[c]{\footnotesize Info-\\\footnotesize GAN} 
& \makecell[c]{\footnotesize BE-\\\footnotesize GAN} 
& \makecell[c]{\footnotesize Cram-\\\footnotesize GAN} 
& \makecell[c]{\footnotesize Att-\\\footnotesize GAN} 
& \makecell[c]{\footnotesize MMD-\\\footnotesize GAN} 
& \makecell[c]{\footnotesize Rel-\\\footnotesize GAN} 
& \makecell[c]{\footnotesize S3-\\\footnotesize GAN} 
& \makecell[c]{\footnotesize SNG-\\\footnotesize GAN} 
& \makecell[c]{\footnotesize STG-\\\footnotesize GAN} 
& \makecell[c]{\footnotesize Pro-\\\footnotesize GAN} 
& \makecell[c]{\footnotesize Style-\\\footnotesize GAN} 
& \makecell[c]{\footnotesize Style-\\\footnotesize GAN2} 
& \makecell[c]{\footnotesize Big-\\\footnotesize GAN} 
& \makecell[c]{\footnotesize Cycle-\\\footnotesize GAN} 
& \makecell[c]{\footnotesize Star-\\\footnotesize GAN} 
& \makecell[c]{\footnotesize Gau-\\\footnotesize GAN} 
& \makecell[c]{\footnotesize Deep-\\\footnotesize fake} 
& \\

\midrule
\footnotesize \href{https://openaccess.thecvf.com/content_ICCV_2019/papers/Rossler_FaceForensics_Learning_to_Detect_Manipulated_Facial_Images_ICCV_2019_paper.pdf}{FF++\dag (2019)}     & \footnotesize 68.9 & \footnotesize 29.9 & \footnotesize 82.0 & \footnotesize 63.3 & \footnotesize 80.4 & \footnotesize 67.2 & \footnotesize 55.5 & \footnotesize 75.4 & \footnotesize 82.0 & \footnotesize 93.0 & \footnotesize 61.1 & \footnotesize 59.8 & \footnotesize 57.9 & \footnotesize 80.1 & \footnotesize 78.6 & \footnotesize 67.3 & \footnotesize 51.9 & \footnotesize 67.9\\

\footnotesize \href{https://arxiv.org/pdf/2007.09355}{F3Net\dag (2020)}      & \footnotesize 62.0 & \footnotesize 43.4 & \footnotesize 65.8 & \footnotesize 53.2 & \footnotesize 64.1 & \footnotesize 56.7 & \footnotesize 55.4 & \footnotesize 58.8 & \footnotesize 67.7 & \footnotesize 92.5 & \footnotesize 76.6 & \footnotesize 62.3 & \footnotesize 56.8 & \footnotesize 60.5 & \footnotesize 71.0 & \footnotesize 71.3 & \footnotesize 51.1 & \footnotesize 63.4\\

\footnotesize \href{https://openaccess.thecvf.com/content/CVPR2021/papers/Zhao_Multi-Attentional_Deepfake_Detection_CVPR_2021_paper.pdf}{MAT\dag (2021)}       & \footnotesize 52.2 & \footnotesize 49.3 & \footnotesize 62.5 & \footnotesize 50.6 & \footnotesize 60.3 & \footnotesize 51.7 & \footnotesize 53.3 & \footnotesize 53.9 & \footnotesize 58.6 & \footnotesize 92.2 & \footnotesize 54.4 & \footnotesize 54.9 & \footnotesize 54.0 & \footnotesize 76.5 & \footnotesize 59.4 & \footnotesize 68.4 & \footnotesize 51.0 & \footnotesize 59.4\\

\footnotesize \href{https://openaccess.thecvf.com/content/CVPR2022/papers/Shiohara_Detecting_Deepfakes_With_Self-Blended_Images_CVPR_2022_paper.pdf}{SBI\dag (2022)}   & \footnotesize 61.3 & \footnotesize 57.4 & \footnotesize 74.8 & \footnotesize 50.3 & \footnotesize 67.5 & \footnotesize 54.6 & \footnotesize 61.5 & \footnotesize 53.2 & \footnotesize 57.1 & \footnotesize 95.9 & \footnotesize 57.2 & \footnotesize 52.9 & \footnotesize 55.4 & \footnotesize 78.3 & \footnotesize 59.3 & \footnotesize 74.6 & \footnotesize 50.7 & \footnotesize 62.6\\

\footnotesize\href{https://openaccess.thecvf.com/content/ICCV2023/papers/Le_Quality-Agnostic_Deepfake_Detection_with_Intra-model_Collaborative_Learning_ICCV_2023_paper.pdf}{QAD\dag (2023)}    & \footnotesize 76.7 & \footnotesize 46.4 & \footnotesize 79.6 & \footnotesize 68.5 & \footnotesize 77.1 & \footnotesize 73.6 & \footnotesize 58.3 & \footnotesize 76.3 & \footnotesize 81.0 & \footnotesize 90.2 & \footnotesize 65.3 & \footnotesize 71.3 & \footnotesize 64.6 & \footnotesize 81.8 & \footnotesize 77.1 & \footnotesize 66.7 & \footnotesize 55.1 & \footnotesize 71.0\\

\footnotesize \href{https://openaccess.thecvf.com/content/CVPR2023/papers/Ojha_Towards_Universal_Fake_Image_Detectors_That_Generalize_Across_Generative_Models_CVPR_2023_paper.pdf}{Ojha\ddag (2023)}                 & \footnotesize 71.5 & \footnotesize 56.2 & \footnotesize 70.4 & \footnotesize 60.0 & \footnotesize 69.5 & \footnotesize 64.1 & \footnotesize 72.9 & \footnotesize 64.8 & \footnotesize 61.2 & \footnotesize 90.2 & \footnotesize 67.0 & \footnotesize 63.6 & \footnotesize 68.4 & \footnotesize 85.3 & \footnotesize 76.3 & \footnotesize 87.8& \footnotesize 61.8& \footnotesize 69.9\\

\footnotesize \href{https://openaccess.thecvf.com/content/CVPR2024/papers/Tan_Rethinking_the_Up-Sampling_Operations_in_CNN-based_Generative_Network_for_Generalizable_CVPR_2024_paper.pdf}{NPR\ddag (2024)} & \footnotesize62.7 & \footnotesize40.6 & \footnotesize57.9 & \footnotesize63.8 & \footnotesize56.7 & \footnotesize70.7 & \footnotesize58.5 & \footnotesize58.5 & \footnotesize58.3 & \footnotesize84.9 & \footnotesize68.7 & \footnotesize71.4 & \footnotesize68.2 & \footnotesize72.6 & \footnotesize69.1 & \footnotesize61.0 & \footnotesize55.5 & \footnotesize63.8\\

\footnotesize \href{https://openaccess.thecvf.com/content/CVPR2024/papers/Liu_Forgery-aware_Adaptive_Transformer_for_Generalizable_Synthetic_Image_Detection_CVPR_2024_paper.pdf}{FaT\ddag (2024)} & \footnotesize62.7 & \footnotesize51.9 & \footnotesize64.8 & \footnotesize62.2 & \footnotesize63.5 & \footnotesize67.4 & \footnotesize54.6 & \footnotesize64.4 & \footnotesize64.4 & \footnotesize85.5 & \footnotesize59.8 & \footnotesize60.9 & \footnotesize60.2 & \footnotesize73.6 & \footnotesize72.0 & \footnotesize59.3 & \footnotesize50.4 & \footnotesize63.5\\

\midrule
\footnotesize \href{https://doi.org/10.1609/aaai.v39i1.32063}{ODDN\dag (2025)}     & \footnotesize 80.4 & \footnotesize 35.1 & \footnotesize 81.0 & \footnotesize 68.7 & \footnotesize 78.2 & \footnotesize 74.5 & \footnotesize 62.2 & \footnotesize 77.5 & \footnotesize 81.7  & \footnotesize 91.7 & \footnotesize 69.2 & \footnotesize 70.4 & \footnotesize 68.0 & \footnotesize 78.8 & \footnotesize 73.4 & \footnotesize 73.8 & \footnotesize 55.3 & \footnotesize \underline{72.1}\\

\rowcolor{gray!20}
\footnotesize PLADA(\textbf{ours})  & \footnotesize 77.8 & \footnotesize 74.2 & \footnotesize 80.8 & \footnotesize 69.7 & \footnotesize 79.4 & \footnotesize 69.0 & \footnotesize 75.4 & \footnotesize 76.4 & \footnotesize 78.1 & \footnotesize 93.3 & \footnotesize 69.9 & \footnotesize 64.3 & \footnotesize 71.8 & \footnotesize 84.4 & \footnotesize 84.3 & \footnotesize 90.0 & \footnotesize 75.3 & \textbf{\footnotesize 77.4}\\

\bottomrule
    \end{tabular}}
\end{table*}

\begin{table*}[!t]
\scriptsize
\centering
\caption{The \textbf{Quality-Agnostic} Experimental Results under a \textbf{4-Class Training Data Setting} on the \textit{DiffusionForensics} Dataset. The marks are consistent with those in Table 1.}
\label{4class_8DMs_Quality_agnostic}

\setlength{\tabcolsep}{0.58mm}{
\begin{tabular}{l | c c | c c | c c | c c | c c | c c | c c | c c | c c}
\toprule
 \multirow{2}{*}{\footnotesize Methods} 
& \multicolumn{2}{c|}{\footnotesize DALLE} 
& \multicolumn{2}{c|}{\footnotesize Glide\_100\_10}
& \multicolumn{2}{c|}{\footnotesize Glide\_100\_27}
& \multicolumn{2}{c|}{\footnotesize Glide\_50\_27}
& \multicolumn{2}{c|}{\footnotesize ADM}
& \multicolumn{2}{c|}{\footnotesize LDM\_100}
& \multicolumn{2}{c|}{\footnotesize LDM\_200}
& \multicolumn{2}{c|}{\footnotesize LDM\_200\_cfg}
& \multicolumn{2}{c}{\footnotesize Mean} \\

\cmidrule{2-19}
& \multicolumn{2}{c|}{\footnotesize Acc. \ \ \footnotesize A.P.}  
& \multicolumn{2}{c|}{\footnotesize Acc. \ \ \footnotesize A.P.}   
& \multicolumn{2}{c|}{\footnotesize Acc. \ \ \footnotesize A.P.}  
& \multicolumn{2}{c|}{\footnotesize Acc. \ \ \footnotesize A.P.} 
& \multicolumn{2}{c|}{\footnotesize Acc. \ \ \footnotesize A.P.}  
& \multicolumn{2}{c|}{\footnotesize Acc. \ \ \footnotesize A.P.}   
& \multicolumn{2}{c|}{\footnotesize Acc. \ \ \footnotesize A.P.}  
& \multicolumn{2}{c|}{\footnotesize Acc. \ \ \footnotesize A.P.} 
& \multicolumn{2}{c}{\footnotesize Acc. \ \ \footnotesize A.P.} \\

\midrule

\footnotesize \href{https://openaccess.thecvf.com/content_ICCV_2019/papers/Rossler_FaceForensics_Learning_to_Detect_Manipulated_Facial_Images_ICCV_2019_paper.pdf}{FF++\ddag (2019)} 
& \multicolumn{2}{c|}{\footnotesize 53.5 \ \ \footnotesize 66.9}  
& \multicolumn{2}{c|}{\footnotesize 52.5 \ \ \footnotesize 63.8}   
& \multicolumn{2}{c|}{\footnotesize 53.7 \ \ \footnotesize 65.9}  
& \multicolumn{2}{c|}{\footnotesize 53.9 \ \ \footnotesize 64.9} 
& \multicolumn{2}{c|}{\footnotesize 55.7 \ \ \footnotesize 67.6}  
& \multicolumn{2}{c|}{\footnotesize 55.9 \ \ \footnotesize 74.0}   
& \multicolumn{2}{c|}{\footnotesize 54.7 \ \ \footnotesize 72.3}  
&\multicolumn{2}{c|}{\footnotesize 53.2 \ \ \footnotesize 65.3} 
& \multicolumn{2}{c}{\footnotesize 54.1 \ \ \footnotesize 67.6} \\

\footnotesize \href{https://arxiv.org/pdf/2007.09355}{F3Net\ddag (2020)} 
& \multicolumn{2}{c|}{\footnotesize 53.8 \ \ \footnotesize 59.5}  
& \multicolumn{2}{c|}{\footnotesize 54.0 \ \ \footnotesize 60.2}   
& \multicolumn{2}{c|}{\footnotesize 53.8 \ \ \footnotesize 60.9}  
& \multicolumn{2}{c|}{\footnotesize 54.9 \ \ \footnotesize 61.9} 
& \multicolumn{2}{c|}{\footnotesize 55.1 \ \ \footnotesize 61.3}  
& \multicolumn{2}{c|}{\footnotesize 53.6 \ \ \footnotesize 59.5}   
& \multicolumn{2}{c|}{\footnotesize 52.7 \ \ \footnotesize 57.7}  
&\multicolumn{2}{c|}{\footnotesize 51.0 \ \ \footnotesize 53.4} 
& \multicolumn{2}{c}{\footnotesize 53.6 \ \ \footnotesize 59.2} \\

\footnotesize \href{https://openaccess.thecvf.com/content/CVPR2021/papers/Zhao_Multi-Attentional_Deepfake_Detection_CVPR_2021_paper.pdf}{MAT\ddag (2021)} 
& \multicolumn{2}{c|}{\footnotesize 50.3 \ \ \footnotesize 45.8}  
& \multicolumn{2}{c|}{\footnotesize 49.8 \ \ \footnotesize 45.0}   
& \multicolumn{2}{c|}{\footnotesize 50.8 \ \ \footnotesize 45.0}  
& \multicolumn{2}{c|}{\footnotesize 50.0 \ \ \footnotesize 43.5} 
& \multicolumn{2}{c|}{\footnotesize 48.2 \ \ \footnotesize 39.3}  
& \multicolumn{2}{c|}{\footnotesize 51.0 \ \ \footnotesize 43.5}   
& \multicolumn{2}{c|}{\footnotesize 50.6 \ \ \footnotesize 44.2}  
&\multicolumn{2}{c|}{\footnotesize 50.0 \ \ \footnotesize 47.5}
& \multicolumn{2}{c}{\footnotesize 50.1 \ \ \footnotesize 44.2} \\

\footnotesize \href{https://openaccess.thecvf.com/content/CVPR2022/papers/Shiohara_Detecting_Deepfakes_With_Self-Blended_Images_CVPR_2022_paper.pdf}{SBI\ddag (2022)} 
& \multicolumn{2}{c|}{\footnotesize 52.7 \ \ \footnotesize 57.3}  
& \multicolumn{2}{c|}{\footnotesize 53.9 \ \ \footnotesize 63.7}   
& \multicolumn{2}{c|}{\footnotesize 54.5 \ \ \footnotesize 63.8}  
& \multicolumn{2}{c|}{\footnotesize 54.4 \ \ \footnotesize 64.8} 
& \multicolumn{2}{c|}{\footnotesize 54.8 \ \ \footnotesize 63.7}  
& \multicolumn{2}{c|}{\footnotesize 57.2 \ \ \footnotesize 70.8}   
& \multicolumn{2}{c|}{\footnotesize 57.1 \ \ \footnotesize 70.3}  
&\multicolumn{2}{c|}{\footnotesize 52.0 \ \ \footnotesize 56.8} 
& \multicolumn{2}{c}{\footnotesize 54.5 \ \ \footnotesize 63.9} \\

\footnotesize \href{https://openaccess.thecvf.com/content/ICCV2023/papers/Le_Quality-Agnostic_Deepfake_Detection_with_Intra-model_Collaborative_Learning_ICCV_2023_paper.pdf}{QAD\ddag (2023)} 
& \multicolumn{2}{c|}{\footnotesize 52.3 \ \ \footnotesize 55.8}  
& \multicolumn{2}{c|}{\footnotesize 59.6 \ \ \footnotesize 67.6}   
& \multicolumn{2}{c|}{\footnotesize 59.7 \ \ \footnotesize 67.9}  
& \multicolumn{2}{c|}{\footnotesize 61.0 \ \ \footnotesize 69.8} 
& \multicolumn{2}{c|}{\footnotesize 56.5 \ \ \footnotesize 58.8}  
& \multicolumn{2}{c|}{\footnotesize 60.1 \ \ \footnotesize 67.9}   
& \multicolumn{2}{c|}{\footnotesize 60.6 \ \ \footnotesize 68.4}  
& \multicolumn{2}{c|}{\footnotesize 49.7 \ \ \footnotesize 50.5} 
& \multicolumn{2}{c}{\footnotesize 58.8 \ \ \footnotesize 63.3} \\

\footnotesize \href{https://openaccess.thecvf.com/content/CVPR2023/papers/Ojha_Towards_Universal_Fake_Image_Detectors_That_Generalize_Across_Generative_Models_CVPR_2023_paper.pdf}{Ojha\ddag (2023)} 
& \multicolumn{2}{c|}{\footnotesize 62.7 \ \ \footnotesize 69.6}  
& \multicolumn{2}{c|}{\footnotesize 75.3 \ \ \footnotesize 84.2}   
& \multicolumn{2}{c|}{\footnotesize 75.8 \ \ \footnotesize 83.2}  
& \multicolumn{2}{c|}{\footnotesize 76.6 \ \ \footnotesize 85.0} 
& \multicolumn{2}{c|}{\footnotesize 65.7 \ \ \footnotesize 70.4}  
& \multicolumn{2}{c|}{\footnotesize 77.0 \ \ \footnotesize 84.5}   
& \multicolumn{2}{c|}{\footnotesize 77.8 \ \ \footnotesize 84.8}  
& \multicolumn{2}{c|}{\footnotesize 60.1 \ \ \footnotesize 64.4} 
& \multicolumn{2}{c}{\footnotesize \textbf{71.4} \ \ \footnotesize \underline{78.3}} \\

\footnotesize \href{https://openaccess.thecvf.com/content/CVPR2024/papers/Tan_Rethinking_the_Up-Sampling_Operations_in_CNN-based_Generative_Network_for_Generalizable_CVPR_2024_paper.pdf}{NPR\ddag (2024)}  
& \multicolumn{2}{c|}{\footnotesize 63.2 \ \ \footnotesize 67.3}  
& \multicolumn{2}{c|}{\footnotesize 61.9 \ \ \footnotesize 64.1}   
& \multicolumn{2}{c|}{\footnotesize 61.2 \ \ \footnotesize 61.8}  
& \multicolumn{2}{c|}{\footnotesize 61.0 \ \ \footnotesize 63.6} 
& \multicolumn{2}{c|}{\footnotesize 76.9 \ \ \footnotesize 77.8}  
& \multicolumn{2}{c|}{\footnotesize 67.0 \ \ \footnotesize 68.1}   
& \multicolumn{2}{c|}{\footnotesize 64.8 \ \ \footnotesize 68.0}  
& \multicolumn{2}{c|}{\footnotesize 63.8 \ \ \footnotesize 67.4} 
& \multicolumn{2}{c}{\footnotesize 65.0 \ \ \footnotesize 67.3} \\

\footnotesize \href{https://openaccess.thecvf.com/content/CVPR2024/papers/Liu_Forgery-aware_Adaptive_Transformer_for_Generalizable_Synthetic_Image_Detection_CVPR_2024_paper.pdf}{FaT\ddag (2024)} 
& \multicolumn{2}{c|}{\footnotesize 58.6 \ \ \footnotesize 56.1}  
& \multicolumn{2}{c|}{\footnotesize 55.4 \ \ \footnotesize 53.5}   
& \multicolumn{2}{c|}{\footnotesize 54.6 \ \ \footnotesize 53.0}  
& \multicolumn{2}{c|}{\footnotesize 55.5 \ \ \footnotesize 53.6} 
& \multicolumn{2}{c|}{\footnotesize 52.9 \ \ \footnotesize 51.7}  
& \multicolumn{2}{c|}{\footnotesize 53.9 \ \ \footnotesize 52.2}   
& \multicolumn{2}{c|}{\footnotesize 53.0 \ \ \footnotesize 51.8}  
& \multicolumn{2}{c|}{\footnotesize 54.8 \ \ \footnotesize 53.1} 
& \multicolumn{2}{c}{\footnotesize 54.8 \ \ \footnotesize 53.1} \\

\midrule
\footnotesize \href{https://doi.org/10.1609/aaai.v39i1.32063}{ODDN\dag (2025)} 
& \multicolumn{2}{c|}{\footnotesize 58.6 \ \ \footnotesize 61.9}  
& \multicolumn{2}{c|}{\footnotesize 55.4 \ \ \footnotesize 57.1}   
& \multicolumn{2}{c|}{\footnotesize 55.7 \ \ \footnotesize 55.9}  
& \multicolumn{2}{c|}{\footnotesize 55.1 \ \ \footnotesize 56.4} 
& \multicolumn{2}{c|}{\footnotesize 59.9 \ \ \footnotesize 60.6}  
& \multicolumn{2}{c|}{\footnotesize 57.7 \ \ \footnotesize 59.4}   
& \multicolumn{2}{c|}{\footnotesize 55.9 \ \ \footnotesize 57.9}  
& \multicolumn{2}{c|}{\footnotesize 53.8 \ \ \footnotesize 55.6} 
& \multicolumn{2}{c}{\footnotesize 56.5 \ \ \footnotesize 58.1} \\

\rowcolor{gray!20}
\footnotesize PLADA(\textbf{ours}) 
& \multicolumn{2}{c|}{\footnotesize 66.5 \ \ \footnotesize 80.3}  
& \multicolumn{2}{c|}{\footnotesize 67.9 \ \ \footnotesize 83.5}   
& \multicolumn{2}{c|}{\footnotesize 69.3 \ \ \footnotesize 84.3}  
& \multicolumn{2}{c|}{\footnotesize 69.9 \ \ \footnotesize 83.3} 
& \multicolumn{2}{c|}{\footnotesize 67.1 \ \ \footnotesize 82.4}  
& \multicolumn{2}{c|}{\footnotesize 85.2 \ \ \footnotesize 94.2}   
& \multicolumn{2}{c|}{\footnotesize 85.0 \ \ \footnotesize 93.7}  
& \multicolumn{2}{c|}{\footnotesize 60.1 \ \ \footnotesize 73.4} 
& \multicolumn{2}{c}{\footnotesize \textbf{71.4} \ \ \footnotesize \textbf{84.4}} \\

\bottomrule
\end{tabular}}
\end{table*}

Overall, the quality-aware results demonstrate that the proposed framework exhibits resilience to compressed images. Despite the limitations posed by scarce paired data, it consistently delivers superior performance across various GANs. Compared to baseline methods, the proposed approach achieves the most competitive or  highest average accuracy, highlighting its effectiveness in distinguishing deepfakes while remaining unaffected by misleading compression artifacts. This robustness underscores its potential, particularly in detecting DM-generated content disseminated on OSNs.

\subsection{Quality-agnostic Evaluation}
In more complex and realistic scenario, deepfake images may be uploaded to and downloaded from OSNs repeatedly. Consequently, it becomes challenging to identify the degree of compression for an image due to the superimposition of these compression operations. And with a reduction in the amount of paired data, it is worthwhile to analyze whether the PLADA can consistently maintain excellent performance. Therefore, we propose a quality-agnostic experiment. Specifically, we sample 10\%, 20\%, 30\%, and 50\% of the data to form paired data, with quantization parameters sampled from the range $[30,100]$ for each image. Subsequently, detectors are trained on each dataset group and then tested. During the inference phase, test images are also compressed with random sampled quantization parameters.

\begin{figure*}[!t]
    \centering
    \setlength{\abovecaptionskip}{-0.4cm}
    \includegraphics[width=\textwidth]{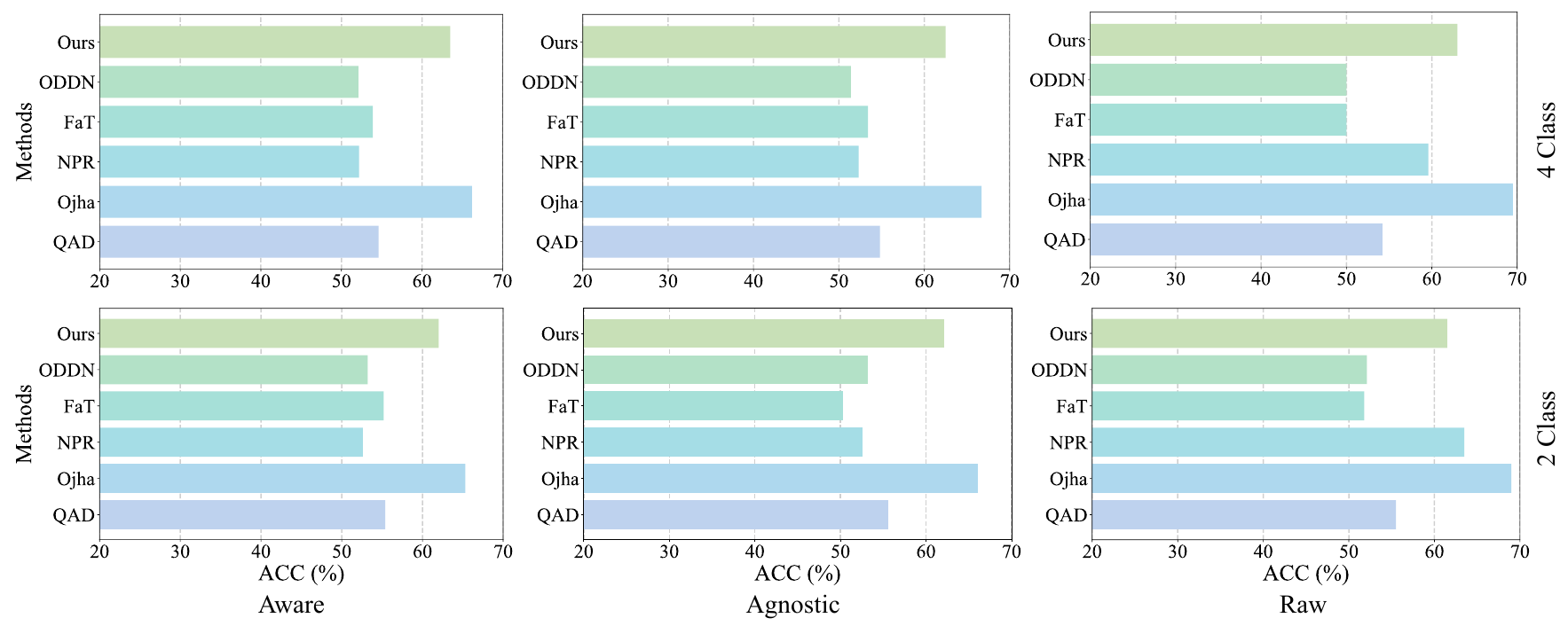} %
    \caption{The \textbf{Quality-Agnostic}, \textbf{Quality-Aware}, and \textbf{Raw Image} Experimental Results on the \textit{DiffusionForensics} Dataset, Evaluated under Both \textbf{4-Class} and \textbf{2-Class} Training Data Settings.}
    \label{5DMs}
\end{figure*}

\begin{figure}[t!]
  \centering
   \includegraphics[scale=0.4]{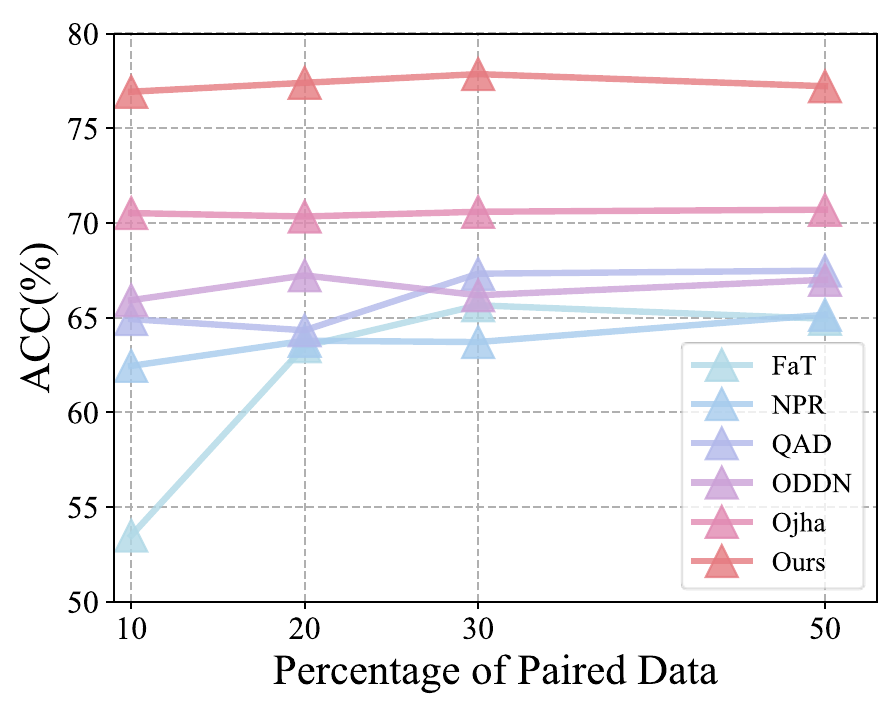}
   \caption{Evaluation in a 4-Class Training Data Setting on the \textit{ForenSynths and \textit{GANGen-Detection} Datasets, Using \textbf{Various Percentages of Paired Data.}} The evaluation was conducted with varying amounts of paired data. The results demonstrate that PLADA exhibits not only robust resilience to compressed images but also remarkable insensitivity to the quantity of paired data, a highly advantageous trait for real-world applications.}
   \label{quality-agnostic}
\end{figure}

The comparative results are presented in Figure \ref{quality-agnostic}. It is evident that our proposed PLADA framework delivers outstanding performance across different amounts of paired data,  maintaining at least a 7\% margin over other methods. Besides, PLADA exhibits remarkable stability while retaining a strong ability to distinguish between real and fake images. Similarly, Ojha's method also utilizes CLIP, but with an simple additional linear projection. Despite exhibiting comparable steadiness, it lags behind PLADA in overall performance. 
Recently, ODDN claimed to possess strong resistance to compression, even surpassing QAD, a method specifically designed for detecting compressed deepfake images. From Figure \ref{quality-agnostic}, it is evident that ODDN does outperform QAD when the amount paired data is extremely limited. However, as the quantity of paired data increases,  the performance gap between ODDN and QAD narrows, and in some instances, QAD even surpasses ODDN. Nevertheless, both methods remain inferior to PLADA. Given the observed variability in the results, it is clear that NPR, FaT, QAD and ODDN can't consistently achieve stable discrimination of compressed deepfakes under complex conditions.

For a more clear comparison, we present the results of the 4-Class and 2-Class agnostic experiments under the 20\% paired data setting. As shown in Table \ref{4class-17GANs-Quality-agnostic}, PLADA outperforms SOTA and other baselines by a significant margin. PLADA achieves the highest performance with a mean accuracy of 77.4\%, while ODDA ranks second with a mean accuracy of 72.1\%, showing a notable gap exceeding 5\%. Despite claims of superior robustness, other methods still fall short compared to PLADA. QAD, designed for detecting compressed deepfakes, lacks robustness compared to PLADA, as it only facilitates interactions between compressed and raw images, whereas PLADA goes further by actively redirecting the model’s attention away from compression artifacts to deepfake artifacts. In Table \ref{4class_8DMs_Quality_agnostic} and Figure \ref{5DMs}, although PLADA has same accuracy with Ojha, it achieves the highest average precision at 84.4\%, surpassing Ojha by 6.1\%. This finding suggests that PLADA possesses a generalization capability comparable to intact CLIP, while demonstrating enhanced stability. We also conduct quality-agnostic experiments with two classes. As illustrated in  Figure \ref{2class}, PLADA emerges as the dominant approach, encompassing a broader scope. This observation confirms its remarkable resilience, even when operating under the constraints of limited data scale. Turning to Figure \ref{5DMs}, it becomes evident that PLADA outperforms other established baselines, in close proximity to the performance of unaltered CLIP.

\begin{table*}[!t]
\scriptsize
\centering
\caption{The \textbf{Raw Images} Experimental Results under a \textbf{4-Class Training Data Setting} on the \textit{ForenSynths} and \textit{GANGen-Detection} Datasets. The marks are consistent with those in Table 1.}
\label{4class Raw Images}

\setlength{\tabcolsep}{0.48mm}{
\begin{tabular}{l| c c c c c c c c c| c c c c c c c c |c}
\toprule
\multirow{3}{*}{\footnotesize Methods} 
& \multicolumn{9}{c|}{\footnotesize \textit{GANGen-Detection} \citep{chuangchuangtan-GANGen-Detection}} 
& \multicolumn{8}{c|}{\footnotesize \textit{ForenSynths} \citep{wang2020cnn}} 
& \multirow{3}{*}{\makecell[c]{\footnotesize Mean \\ \footnotesize Acc}} \\

\cmidrule(lr){2-10} \cmidrule(lr){11-18}
& \makecell[c]{\footnotesize Info-\\\footnotesize GAN} 
& \makecell[c]{\footnotesize BE-\\\footnotesize GAN} 
& \makecell[c]{\footnotesize Cram-\\\footnotesize GAN} 
& \makecell[c]{\footnotesize Att-\\\footnotesize GAN} 
& \makecell[c]{\footnotesize MMD-\\\footnotesize GAN} 
& \makecell[c]{\footnotesize Rel-\\\footnotesize GAN} 
& \makecell[c]{\footnotesize S3-\\\footnotesize GAN} 
& \makecell[c]{\footnotesize SNG-\\\footnotesize GAN} 
& \makecell[c]{\footnotesize STG-\\\footnotesize GAN} 
& \makecell[c]{\footnotesize Pro-\\\footnotesize GAN} 
& \makecell[c]{\footnotesize Style-\\\footnotesize GAN} 
& \makecell[c]{\footnotesize Style-\\\footnotesize GAN2} 
& \makecell[c]{\footnotesize Big-\\\footnotesize GAN} 
& \makecell[c]{\footnotesize Cycle-\\\footnotesize GAN} 
& \makecell[c]{\footnotesize Star-\\\footnotesize GAN} 
& \makecell[c]{\footnotesize Gau-\\\footnotesize GAN} 
& \makecell[c]{\footnotesize Deep-\\\footnotesize fake} 
& \\

\midrule
\footnotesize \href{https://openaccess.thecvf.com/content_ICCV_2019/papers/Rossler_FaceForensics_Learning_to_Detect_Manipulated_Facial_Images_ICCV_2019_paper.pdf}{FF++\ddag (2019)}   & \footnotesize 80.4 & \footnotesize 73.4 & \footnotesize 91.4 & \footnotesize 65.3 & \footnotesize 88.9 & \footnotesize 78.7 & \footnotesize 59.9 & \footnotesize 86.0 & \footnotesize 79.7 & \footnotesize 97.6 & \footnotesize 85.1 & \footnotesize 90.6 & \footnotesize 71.4 & \footnotesize 79.0 & \footnotesize 88.1 & \footnotesize 68.1 & \footnotesize 58.7 & \footnotesize 79.0\\

\footnotesize \href{https://arxiv.org/pdf/2007.09355}{F3Net\ddag (2020)}   & \footnotesize 80.5 & \footnotesize 79.3 & \footnotesize 89.9 & \footnotesize 63.6 & \footnotesize 87.6 & \footnotesize 77.7 & \footnotesize 57.3 & \footnotesize 82.1 & \footnotesize 81.3 & \footnotesize 97.0 & \footnotesize 80.2 & \footnotesize 85.7 & \footnotesize 68.7 & \footnotesize 76.8 & \footnotesize 90.2 & \footnotesize 72.5 & \footnotesize 56.2 & \footnotesize 78.1\\

\footnotesize \href{https://openaccess.thecvf.com/content/CVPR2021/papers/Zhao_Multi-Attentional_Deepfake_Detection_CVPR_2021_paper.pdf}{MAT\ddag (2021)}   & \footnotesize 48.7 & \footnotesize 50.0 & \footnotesize 44.0 & \footnotesize 59.5 & \footnotesize 45.8 & \footnotesize 48.9 & \footnotesize 48.1 & \footnotesize 49.1 & \footnotesize 48.6 & \footnotesize 11.5 & \footnotesize 47.4 & \footnotesize 48.1 & \footnotesize 47.8 & \footnotesize 27.5 & \footnotesize 44.3 & \footnotesize 38.2 & \footnotesize 49.5 & \footnotesize 43.7\\

\footnotesize \href{https://openaccess.thecvf.com/content/CVPR2022/papers/Shiohara_Detecting_Deepfakes_With_Self-Blended_Images_CVPR_2022_paper.pdf}{SBI\ddag (2022)}   & \footnotesize 58.6 & \footnotesize 50.4 & \footnotesize 74.0 & \footnotesize 50.7 & \footnotesize 70.2 & \footnotesize 54.0 & \footnotesize 55.2 & \footnotesize 55.1 & \footnotesize 59.9 & \footnotesize 96.9 & \footnotesize 59.0 & \footnotesize 57.6 & \footnotesize 54.5 & \footnotesize 83.8 & \footnotesize 60.9 & \footnotesize 70.3 & \footnotesize 50.8 & \footnotesize 62.7\\

\footnotesize \href{https://openaccess.thecvf.com/content/ICCV2023/papers/Le_Quality-Agnostic_Deepfake_Detection_with_Intra-model_Collaborative_Learning_ICCV_2023_paper.pdf}{QAD\ddag (2023)}   & \footnotesize 84.8 & \footnotesize 73.6 & \footnotesize 92.8 & \footnotesize 57.3 & \footnotesize 86.8 & \footnotesize 69.3 & \footnotesize 61.1 & \footnotesize 83.8 & \footnotesize 79.9 & \footnotesize 98.9 & \footnotesize 80.3 & \footnotesize 90.6 & \footnotesize 68.9 & \footnotesize 82.0 & \footnotesize 91.2 & \footnotesize 78.5 & \footnotesize 53.7 & \footnotesize 78.6\\

\footnotesize \href{https://openaccess.thecvf.com/content/CVPR2023/papers/Ojha_Towards_Universal_Fake_Image_Detectors_That_Generalize_Across_Generative_Models_CVPR_2023_paper.pdf}{Ojha\ddag (2023)}  & \footnotesize 81.8 & \footnotesize 60.4 & \footnotesize 81.1 & \footnotesize 65.4 & \footnotesize 79.9 & \footnotesize 67.0 & \footnotesize 78.1 & \footnotesize 71.5 & \footnotesize 65.1 & \footnotesize 93.8 & \footnotesize 69.2 & \footnotesize 63.3 & \footnotesize 73.5 & \footnotesize 89.7 & \footnotesize 79.5 & \footnotesize 93.1& \footnotesize 62.6& \footnotesize 75.2\\

\footnotesize \href{https://openaccess.thecvf.com/content/CVPR2024/papers/Tan_Rethinking_the_Up-Sampling_Operations_in_CNN-based_Generative_Network_for_Generalizable_CVPR_2024_paper.pdf}{NPR\ddag (2024)}  & \footnotesize 83.9 & \footnotesize 66.1 & \footnotesize 89.6 & \footnotesize 84.6 & \footnotesize 86.9 & \footnotesize 91.9 & \footnotesize 67.3 & \footnotesize 88.2 & \footnotesize 85.5 & \footnotesize 98.3 & \footnotesize 92.0 & \footnotesize 93.4 & \footnotesize 72.1 & \footnotesize 76.1 & \footnotesize 93.4 & \footnotesize 64.4 & \footnotesize 61.7& \footnotesize \textbf{82.0}\\

\footnotesize \href{https://openaccess.thecvf.com/content/CVPR2024/papers/Liu_Forgery-aware_Adaptive_Transformer_for_Generalizable_Synthetic_Image_Detection_CVPR_2024_paper.pdf}{FaT\ddag (2024)}  & \footnotesize 74.1 & \footnotesize 64.0 & \footnotesize 75.7 & \footnotesize 77.4 & \footnotesize 75.1 & \footnotesize 78.9 & \footnotesize 63.5 & \footnotesize 74.3 & \footnotesize 76.0 & \footnotesize 97.2 & \footnotesize 78.9 & \footnotesize 82.8 & \footnotesize 73.9 & \footnotesize 80.0 & \footnotesize 73.9 & \footnotesize 73.0& \footnotesize 61.5& \footnotesize 75.4\\

\midrule
\footnotesize \href{https://doi.org/10.1609/aaai.v39i1.32063}{ODDN\ddag (2025)}  & \footnotesize 81.0 & \footnotesize 57.4 & \footnotesize 85.3 & \footnotesize 58.1 & \footnotesize 80.9 & \footnotesize 65.5 & \footnotesize 56.7 & \footnotesize 80.7 & \footnotesize 84.3  & \footnotesize 94.9 & \footnotesize 70.9 & \footnotesize 77.4 & \footnotesize 63.2 & \footnotesize 74.3 & \footnotesize 81.9 & \footnotesize 65.2 & \footnotesize 56.2 & \footnotesize 72.6\\

\rowcolor{gray!20}
\footnotesize PLADA(\textbf{ours}) & \footnotesize 83.6 & \footnotesize 79.0 & \footnotesize 83.7 & \footnotesize 73.0 & \footnotesize 83.9 & \footnotesize 73.2 & \footnotesize 80.3 & \footnotesize 80.9 & \footnotesize 82.9 & \footnotesize 94.7 & \footnotesize 68.4 & \footnotesize 60.8 & \footnotesize 78.8 & \footnotesize 90.4 & \footnotesize 82.9 & \footnotesize 94.1 & \footnotesize 75.0 & \footnotesize \underline{80.4}\\

\bottomrule
    \end{tabular}}
\end{table*}

\begin{table*}[!t]
\scriptsize
\centering
\caption{The \textbf{Raw Images} Experimental Results under a \textbf{4-Class Training Data Setting} on the \textit{Ojha-test} Dataset. The marks are consistent with those in Table 1.}
\label{4class_8DMs_Quality_raw}

\setlength{\tabcolsep}{0.58mm}{
\begin{tabular}{l | c c | c c | c c | c c | c c | c c | c c | c c | c c}
\toprule
 \multirow{2}{*}{\footnotesize Methods} 
& \multicolumn{2}{c|}{\footnotesize DALLE} 
& \multicolumn{2}{c|}{\footnotesize Glide\_100\_10}
& \multicolumn{2}{c|}{\footnotesize Glide\_100\_27}
& \multicolumn{2}{c|}{\footnotesize Glide\_50\_27}
& \multicolumn{2}{c|}{\footnotesize ADM}
& \multicolumn{2}{c|}{\footnotesize LDM\_100}
& \multicolumn{2}{c|}{\footnotesize LDM\_200}
& \multicolumn{2}{c|}{\footnotesize LDM\_200\_cfg}
& \multicolumn{2}{c}{\footnotesize Mean} \\

\cmidrule{2-19}
& \multicolumn{2}{c|}{\footnotesize Acc. \ \ \footnotesize A.P.}  
& \multicolumn{2}{c|}{\footnotesize Acc. \ \ \footnotesize A.P.}   
& \multicolumn{2}{c|}{\footnotesize Acc. \ \ \footnotesize A.P.}  
& \multicolumn{2}{c|}{\footnotesize Acc. \ \ \footnotesize A.P.} 
& \multicolumn{2}{c|}{\footnotesize Acc. \ \ \footnotesize A.P.}  
& \multicolumn{2}{c|}{\footnotesize Acc. \ \ \footnotesize A.P.}   
& \multicolumn{2}{c|}{\footnotesize Acc. \ \ \footnotesize A.P.}  
& \multicolumn{2}{c|}{\footnotesize Acc. \ \ \footnotesize A.P.} 
& \multicolumn{2}{c}{\footnotesize Acc. \ \ \footnotesize A.P.} \\

\midrule

\footnotesize \href{https://openaccess.thecvf.com/content_ICCV_2019/papers/Rossler_FaceForensics_Learning_to_Detect_Manipulated_Facial_Images_ICCV_2019_paper.pdf}{FF++\ddag (2019)} 
& \multicolumn{2}{c|}{\footnotesize 62.9 \ \ \footnotesize 74.5}  
& \multicolumn{2}{c|}{\footnotesize 62.0 \ \ \footnotesize 72.4}   
& \multicolumn{2}{c|}{\footnotesize 60.8 \ \ \footnotesize 70.0}  
& \multicolumn{2}{c|}{\footnotesize 64.6 \ \ \footnotesize 75.6} 
& \multicolumn{2}{c|}{\footnotesize 63.7 \ \ \footnotesize 69.0}  
& \multicolumn{2}{c|}{\footnotesize 59.3 \ \ \footnotesize 71.2}   
& \multicolumn{2}{c|}{\footnotesize 59.7 \ \ \footnotesize 70.7}  
&\multicolumn{2}{c|}{\footnotesize 60.0 \ \ \footnotesize 71.9} 
& \multicolumn{2}{c}{\footnotesize 61.6 \ \ \footnotesize 71.9} \\

\footnotesize \href{https://arxiv.org/pdf/2007.09355}{F3Net\ddag (2020)} 
& \multicolumn{2}{c|}{\footnotesize 59.4 \ \ \footnotesize 68.7}  
& \multicolumn{2}{c|}{\footnotesize 59.8 \ \ \footnotesize 69.3}   
& \multicolumn{2}{c|}{\footnotesize 60.3 \ \ \footnotesize 69.6}  
& \multicolumn{2}{c|}{\footnotesize 61.9 \ \ \footnotesize 72.7} 
& \multicolumn{2}{c|}{\footnotesize 59.8 \ \ \footnotesize 63.7}  
& \multicolumn{2}{c|}{\footnotesize 60.4 \ \ \footnotesize 69.6}   
& \multicolumn{2}{c|}{\footnotesize 59.5 \ \ \footnotesize 68.4}  
&\multicolumn{2}{c|}{\footnotesize 57.3 \ \ \footnotesize 64.5} 
& \multicolumn{2}{c}{\footnotesize 59.8 \ \ \footnotesize 68.3} \\

\footnotesize \href{https://openaccess.thecvf.com/content/CVPR2021/papers/Zhao_Multi-Attentional_Deepfake_Detection_CVPR_2021_paper.pdf}{MAT\ddag (2021)} 
& \multicolumn{2}{c|}{\footnotesize 59.9 \ \ \footnotesize 46.1}  
& \multicolumn{2}{c|}{\footnotesize 58.6 \ \ \footnotesize 44.2}   
& \multicolumn{2}{c|}{\footnotesize 58.7 \ \ \footnotesize 44.4}  
& \multicolumn{2}{c|}{\footnotesize 58.1 \ \ \footnotesize 43.0} 
& \multicolumn{2}{c|}{\footnotesize 44.0 \ \ \footnotesize 38.7}  
& \multicolumn{2}{c|}{\footnotesize 62.1 \ \ \footnotesize 46.2}   
& \multicolumn{2}{c|}{\footnotesize 62.0 \ \ \footnotesize 46.2}  
&\multicolumn{2}{c|}{\footnotesize 61.6 \ \ \footnotesize 48.1}
& \multicolumn{2}{c}{\footnotesize 58.1 \ \ \footnotesize 44.6} \\

\footnotesize \href{https://openaccess.thecvf.com/content/CVPR2022/papers/Shiohara_Detecting_Deepfakes_With_Self-Blended_Images_CVPR_2022_paper.pdf}{SBI\ddag (2022)} 
& \multicolumn{2}{c|}{\footnotesize 52.1 \ \ \footnotesize 59.4}  
& \multicolumn{2}{c|}{\footnotesize 54.7 \ \ \footnotesize 67.4}   
& \multicolumn{2}{c|}{\footnotesize 54.8 \ \ \footnotesize 67.0}  
& \multicolumn{2}{c|}{\footnotesize 55.8 \ \ \footnotesize 69.8} 
& \multicolumn{2}{c|}{\footnotesize 54.6 \ \ \footnotesize 63.7}  
& \multicolumn{2}{c|}{\footnotesize 57.2 \ \ \footnotesize 73.2}   
& \multicolumn{2}{c|}{\footnotesize 56.9 \ \ \footnotesize 73.5}  
&\multicolumn{2}{c|}{\footnotesize 52.3 \ \ \footnotesize 60.3}
& \multicolumn{2}{c}{\footnotesize 54.9 \ \ \footnotesize 66.8} \\

\footnotesize \href{https://openaccess.thecvf.com/content/ICCV2023/papers/Le_Quality-Agnostic_Deepfake_Detection_with_Intra-model_Collaborative_Learning_ICCV_2023_paper.pdf}{QAD\ddag (2023)}
& \multicolumn{2}{c|}{\footnotesize 59.9 \ \ \footnotesize 71.7}  
& \multicolumn{2}{c|}{\footnotesize 69.3 \ \ \footnotesize 83.6}   
& \multicolumn{2}{c|}{\footnotesize 68.2 \ \ \footnotesize 81.3}  
& \multicolumn{2}{c|}{\footnotesize 74.0 \ \ \footnotesize 87.5} 
& \multicolumn{2}{c|}{\footnotesize 62.6 \ \ \footnotesize 65.4}  
& \multicolumn{2}{c|}{\footnotesize 61.0 \ \ \footnotesize 73.7}   
& \multicolumn{2}{c|}{\footnotesize 60.1 \ \ \footnotesize 71.0}  
& \multicolumn{2}{c|}{\footnotesize 62.0 \ \ \footnotesize 73.7} 
& \multicolumn{2}{c}{\footnotesize 64.6 \ \ \footnotesize 76.0} \\

\footnotesize \href{https://openaccess.thecvf.com/content/CVPR2023/papers/Ojha_Towards_Universal_Fake_Image_Detectors_That_Generalize_Across_Generative_Models_CVPR_2023_paper.pdf}{Ojha\ddag (2023)} 
& \multicolumn{2}{c|}{\footnotesize 68.1 \ \ \footnotesize 77.0}  
& \multicolumn{2}{c|}{\footnotesize 78.9 \ \ \footnotesize 87.3}   
& \multicolumn{2}{c|}{\footnotesize 78.3 \ \ \footnotesize 86.8}  
& \multicolumn{2}{c|}{\footnotesize 81.2 \ \ \footnotesize 88.8} 
& \multicolumn{2}{c|}{\footnotesize 70.1 \ \ \footnotesize 75.4}  
& \multicolumn{2}{c|}{\footnotesize 81.7 \ \ \footnotesize 89.6}   
& \multicolumn{2}{c|}{\footnotesize 82.9 \ \ \footnotesize 90.2}  
& \multicolumn{2}{c|}{\footnotesize 64.6 \ \ \footnotesize 71.6} 
& \multicolumn{2}{c}{\footnotesize \textbf{75.7} \ \ \footnotesize 83.4} \\

\footnotesize \href{https://openaccess.thecvf.com/content/CVPR2024/papers/Tan_Rethinking_the_Up-Sampling_Operations_in_CNN-based_Generative_Network_for_Generalizable_CVPR_2024_paper.pdf}{NPR\ddag (2024)}  
& \multicolumn{2}{c|}{\footnotesize 75.7 \ \ \footnotesize 86.4}  
& \multicolumn{2}{c|}{\footnotesize 75.3 \ \ \footnotesize 87.0}   
& \multicolumn{2}{c|}{\footnotesize 74.1 \ \ \footnotesize 85.8}  
& \multicolumn{2}{c|}{\footnotesize 77.3 \ \ \footnotesize 89.0} 
& \multicolumn{2}{c|}{\footnotesize 65.7 \ \ \footnotesize 68.0}  
& \multicolumn{2}{c|}{\footnotesize 79.7 \ \ \footnotesize 89.5}   
& \multicolumn{2}{c|}{\footnotesize 78.9 \ \ \footnotesize 88.6}  
& \multicolumn{2}{c|}{\footnotesize 78.7 \ \ \footnotesize 89.3} 
& \multicolumn{2}{c}{\footnotesize \textbf{75.7} \ \ \footnotesize \underline{85.4}} \\

\footnotesize \href{https://openaccess.thecvf.com/content/CVPR2024/papers/Liu_Forgery-aware_Adaptive_Transformer_for_Generalizable_Synthetic_Image_Detection_CVPR_2024_paper.pdf}{FaT\ddag (2024)} 
& \multicolumn{2}{c|}{\footnotesize 66.5 \ \ \footnotesize 62.3}  
& \multicolumn{2}{c|}{\footnotesize 55.2 \ \ \footnotesize 53.2}   
& \multicolumn{2}{c|}{\footnotesize 54.5 \ \ \footnotesize 52.7}  
& \multicolumn{2}{c|}{\footnotesize 56.2 \ \ \footnotesize 53.9} 
& \multicolumn{2}{c|}{\footnotesize 56.1 \ \ \footnotesize 53.5}  
& \multicolumn{2}{c|}{\footnotesize 64.5 \ \ \footnotesize 60.6}   
& \multicolumn{2}{c|}{\footnotesize 64.4 \ \ \footnotesize 60.5}  
& \multicolumn{2}{c|}{\footnotesize 64.2 \ \ \footnotesize 60.3} 
& \multicolumn{2}{c}{\footnotesize 60.2 \ \ \footnotesize 57.1} \\

\midrule
\footnotesize \href{https://doi.org/10.1609/aaai.v39i1.32063}{ODDN\ddag (2025)} 
& \multicolumn{2}{c|}{\footnotesize 60.6 \ \ \footnotesize 66.1}  
& \multicolumn{2}{c|}{\footnotesize 67.0 \ \ \footnotesize 72.6}   
& \multicolumn{2}{c|}{\footnotesize 65.2 \ \ \footnotesize 70.4}  
& \multicolumn{2}{c|}{\footnotesize 70.6 \ \ \footnotesize 76.4} 
& \multicolumn{2}{c|}{\footnotesize 57.1 \ \ \footnotesize 58.6}  
& \multicolumn{2}{c|}{\footnotesize 58.0 \ \ \footnotesize 60.4}   
& \multicolumn{2}{c|}{\footnotesize 57.2 \ \ \footnotesize 58.1}  
& \multicolumn{2}{c|}{\footnotesize 59.2 \ \ \footnotesize 61.1} 
& \multicolumn{2}{c}{\footnotesize 61.8 \ \ \footnotesize 65.5} \\

\rowcolor{gray!20}
\footnotesize PLADA(\textbf{ours}) 
& \multicolumn{2}{c|}{\footnotesize 70.9 \ \ \footnotesize 85.0}  
& \multicolumn{2}{c|}{\footnotesize 68.1 \ \ \footnotesize 85.3}   
& \multicolumn{2}{c|}{\footnotesize 70.5 \ \ \footnotesize 86.1}  
& \multicolumn{2}{c|}{\footnotesize 70.9 \ \ \footnotesize 86.2} 
& \multicolumn{2}{c|}{\footnotesize 65.3 \ \ \footnotesize 82.8}  
& \multicolumn{2}{c|}{\footnotesize 85.2 \ \ \footnotesize 94.9}   
& \multicolumn{2}{c|}{\footnotesize 84.3 \ \ \footnotesize 94.9}  
& \multicolumn{2}{c|}{\footnotesize 61.2 \ \ \footnotesize 77.3} 
& \multicolumn{2}{c}{\footnotesize \underline{72.0} \ \ \footnotesize \textbf{86.5}} \\

\bottomrule
\end{tabular}}
\end{table*}

In conclusion, PLADA consistently delivers exceptional performance in deepfake detection, regardless of the amount of paired data available. Quality-agnostic experiments further confirm that PLADA can effectively identify subtle correlations, no matter the fixed image quality of the paired data. These strengths are particularly pronounced when compared to QAD, which relies on the Hilbert-Schmidt Independence Criterion and is mainly designed for quality-aware scenarios.

\subsection{Raw Images Evaluation}
While the model excels in detecting compressed deepfakes, we aim to ensure it has outstanding performance in detecting raw deepfakes as well. Therefore, we conduct the raw images experiments in both 4-Class and 2-Class settings to validate the superiority of PLADA. Importantly, the setting was consistent with those used in the 20\% quality-agnostic experiment, however, raw images were employed during the testing phase. 

The results of the 4-Class experiments are presented in Table \ref{4class Raw Images}, Table \ref{4class_8DMs_Quality_raw} and Figure \ref{5DMs}. Although PLADA performs slightly worse than a few of baselines, specifically NPR and Ojha, the gap is negligible when compared to the extent of improvement in detecting compressed images. In other words, this minor decline is acceptable for practical application. Moreover, PLADA achieves the highest AP, indicating that, when considering both precision and recall comprehensively, PLADA actually outperforms the others.

The results of the 2-Class are shown in Figure \ref{2class} and Figure \ref{5DMs}, reflecting the model's generalization capabilities. Concretely, in Figure \ref{2class}, QAD experienced a significant drop in accuracy, declining from 78.6\% in the 4-Class to 70.9\%. Conversely, ODDN exhibited the least performance degradation but still fell short of PLADA. Ojha and FaT, which share the similar backbone as PLADA, lacks a dedicated mechanism to address deceptive compression artifacts, resulting in its shortfall. 

In summary, the experimental results provide evidence that, despite PLADA's specific mechanism designed to defend against compression attacks, it retains sufficient capability to discern raw deepfakes in real-world scenarios.

\begin{figure}[!t]
\centering
\setlength{\abovecaptionskip}{-0.2cm}
   \includegraphics[scale=0.33]{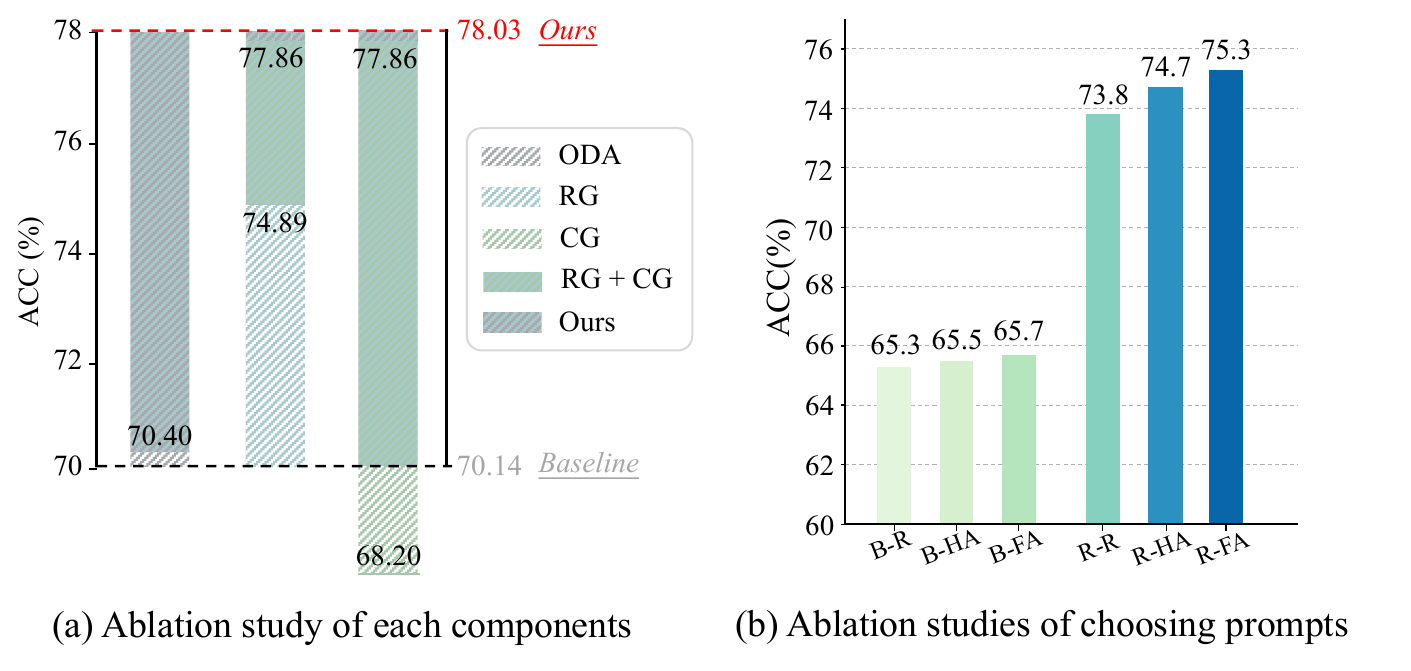}
   \caption{\textbf{Ablation Study of Components and Prompt Selection.} Left (a) illustrates the distinct functions of each component and their interaction results. Right (b) presents the outcomes of various prompt selection strategies. R-FA achieves superior performance by utilizing random sampling for guide prompts during training and averaging the guide pool during inference.}
   \label{components and choosing prompt}
\end{figure}

\begin{figure}
    \centering
    \includegraphics[width=0.65\linewidth]{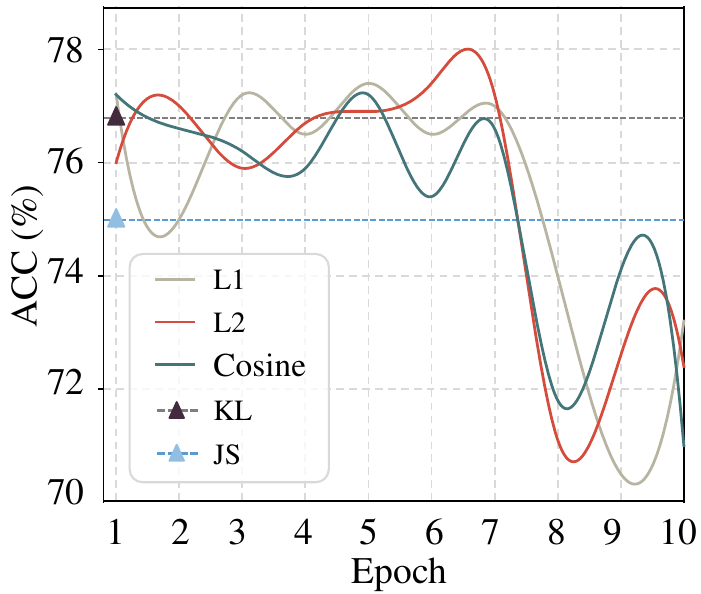}
    \caption{\textbf{Accuracy Curves with Different Distance Objective Functions.} L2 demonstrates the most stable and superior performance compared to the other functions.}
    \label{Distance Loss}
\end{figure}

\subsection{Ablation Study}
\subsubsection{Model Components}
The ablation study in Figure \ref{components and choosing prompt} (a) evaluates the impact of each component on the performance of the baseline method. We select Ojha’s work \citep{ojha2023towards} as the baseline due to its shared backbone architecture with ours. The baseline achieves 70.14\% accuracy, highlighting its insufficient robustness against compressed deepfakes.

When ODA is applied, the accuracy increases to 70.40\%. This improvement is attributed to ODA’s ability to explore subtle information in both unpaired and paired data. B2E incorporates two specialized attention-shifting techniques: RG and CG. We analyze the effect of each technique separately. When only RG is employed, the accuracy significantly rises to 74.89\%, indicating that a deep stack of RG effectively captures guidance signals and redirects attention to deepfake artifacts. Conversely, using only CG introduces tremendous adversarial effects to the model. Because CG refines RG’s output and accelerates attention shifting in a fine-grained manner, it performs suboptimally when used in isolation. Notably, combining RG and CG into B2E significantly boosts accuracy to 77.86\%. This strong performance indicates that RG and CG complement each other, working together to enhance the model’s effectiveness. The complete PLADA framework achieves an accuracy of 78.03\%, maintaining a substantial margin over the naive baseline. 

In summary, each component of PLADA plays a distinct role, contributing to the effectiveness of the framework. The ablation study provides insights into the importance of each component and their interactions within PLADA.

\subsubsection{Distance-Based Loss Function}
By augmenting the distance between cluster centers, ODA facilitates interactions among diverse compression types of both real and deepfake images. Consequently, it is imperative to ascertain the method for computing this distance that maximizes effectiveness. In light of this inquiry, we conducted experiments pertinent to it. We selected three conventional methods for calculating distance: Manhattan distance (L1), Euclidean distance (L2), and Cosine Distance (Cosine). Given that the cluster centers are vectors and can be considered as distributions, we also incorporated two additional methods for computing distribution similarity: Kullback-Leibler Divergence (KL) and Jensen–Shannon Divergence (JS).

As depicted in Figure \ref{Distance Loss}, the three approaches exhibit fluctuations prior to the eighth epoch and demonstrate overfitting thereafter. Regrettably, before KL and JS completed their training process, they led to model collapse, yielding only results from the first epoch. From a broader perspective, L2 achieves the minimal magnitude of fluctuation and peaks in performance. Due to its lack of magnitude representation, Cosine Distance only captures directional distance, which limits its ability to provide a more comprehensive distance representation. Regarding L1, while offering a straightforward computation, it sacrifices considerable distance encoding space. Therefore, L2 emerges as a competitive candidate, and we have selected it as our final approach.

\begin{figure*}[!t]
\centering
\setlength{\abovecaptionskip}{-0.2cm}
\includegraphics[width=\textwidth]{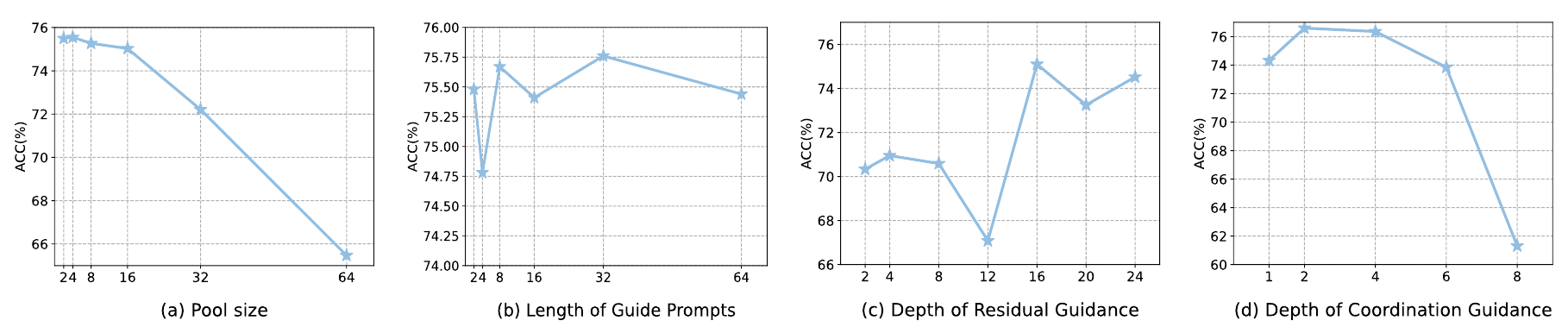} 
\caption{\textbf{Ablation Study of B2E.} We report the accuracy of various configurations to highlight the role of each component and conduct hyperparameter experiments to identify the optimal settings. The results suggest that a smaller pool size, moderately long guide prompts, deep residual guidance, and shallow coordination guidance form the preferred configuration.}
\label{ablation}
\end{figure*}

\begin{figure*}[!t]
\centering
\setlength{\abovecaptionskip}{-0.2cm}
\includegraphics[width=\textwidth]{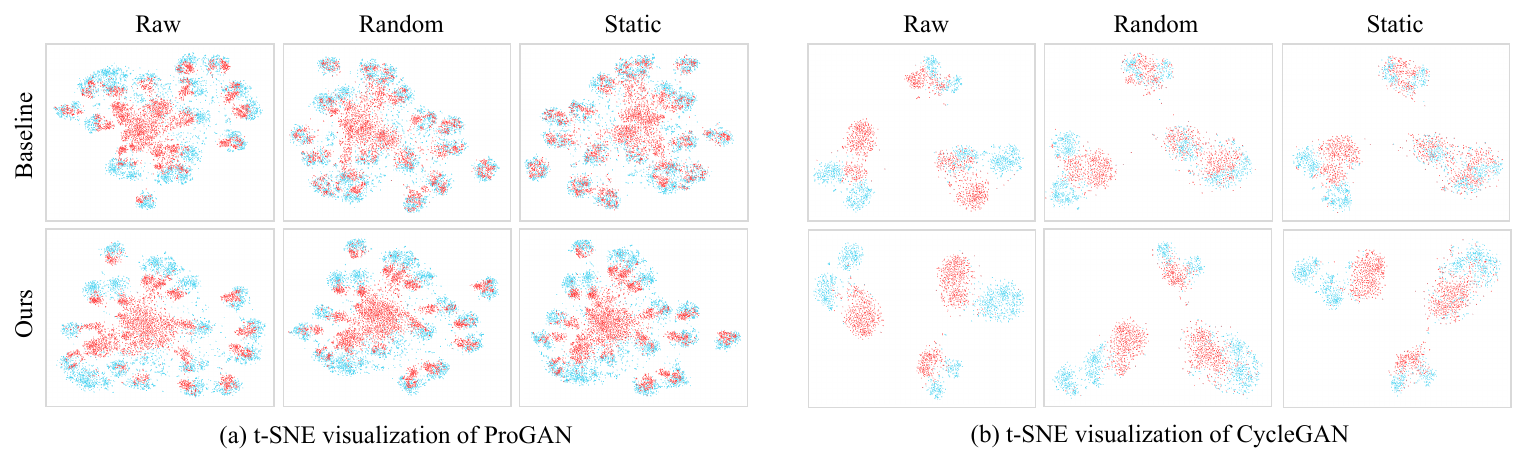} 
\caption{\textbf{The t-SNE Visualization.} Datasets are preprocessed in three forms: original raw, random compression, and static compression. The left panel displays results obtained using ProGAN, while the right panel shows outcomes generated by CycleGAN. Comparisons demonstrate that our PLADA method exhibits robust resilience to compressed deepfakes.}
\label{feature visual}
\end{figure*}

\subsubsection{Selection of Guide Prompts}
As a specialized guidance storage mechanism and a crucial component of the attention shifting process, guide prompts have multiple options for their selection. Specifically, these include Random Sampling (R), Beam Search (B), the Average of all prompts (FA), and the Average of half of prompts (HA). For the training phase, only Random Sampling and Beam Search are employed, as we have observed that FA and HA  negatively impact the training process.  Conversely, for the inference phase, three selection methods are utilized, including R, FA and HA. We exclude Beam Search, since it can be considered as a mathematical extension of both FA and HA. The abbreviation B-R signifies the use of Beam Search during the training phase and Random Sampling during inference. 

Figure \ref{components and choosing prompt} (b) illustrates the impact of various prompt selection configurations. Notably, Beam Search negatively affects detection accuracy, regardless of the inference method used. However, employing Random Sampling during training significantly improves performance. We hypothesize that randomness can mitigate the inclusion of excessive low-quality information and disruptive objective class messages within guide prompts, thereby enhancing their effectiveness. For inference, the results indicate that FA outperforms both R and HA, and this trend is consistent across different training conditions.  To fully exploit the potential of guide prompts, we recommend the R-FA configuration, as it yields the best results according to our experimental findings. 

\subsubsection{Size of Prompt Pool and Length of Guide Prompts}
The pool size determines the number of guide prompts available for selection in each intermediate layer, while the length of these guide prompts dictates the capacity for storing and utilizing attention-shifting signals. Figure \ref{ablation} (a) illustrates the accuracy trend as the pool size increases. Initially, accuracy slightly improves as the pool size increases from 2 to 4. But subsequently, it decreases steadily with further increases in pool size. We hypothesize that a larger pool size may reduce the ability of individual prompts to retain sufficient guidance signals for directing attention shifting, ultimately leading to degraded performance. As illustrated in Figure \ref{ablation} (b), accuracy initially decreases slightly as the length of guide prompt increases and then fluctuates between 75.75\% and 75.25\%. The best performance is achieved when the prompt length is set to 32, indicating that this length is sufficient for guide prompts to encode diverse information effectively. 

\subsubsection{Depth of Residual Guidance Layer}
Residual Guidance is a variant of MSA that incorporates special residual knowledge injection. Similar to traditional MSA, it achieves deep attention through the stacking of multiple layers. In this section,we analyze the optimal number of MSA layers to replace with RG for maximum effectiveness. To isolate the impact of CG, we exclude CG from all layers in our experiments. The result is presented in Figure \ref{ablation} (c). The detector demonstrates the most competitive performance when incorporating 16 layers of RG. And as the number of stacked RG layers increases, the discrimination ability of the model initially rises until it reaches 4 layers, followed by a significant drop from 4 to 12 layers. However, a subsequent continuous increase trend in accuracy suggests that RG successfully directs the model's attention to forgery cues by injecting guidance signals. This trend indicates that RG is well-suited for deep stacking, though its ability to enhance the robustness of the detector may be limited when used in single layer.

\subsubsection{Depth of Coordination Guidance Layer}
Figure \ref{ablation} (d) illustrates the performance under different settings of CG depth.  According to aforementioned model structure, CG must be integrated with either RG or conventional MSA as a precursor. Consequently, we predetermined 24 layers of RG for this experiment. CG is designed to shift attention across both global and local contexts, enabling faster convergence on target objectives. As illustrated in the figure,  performance gradually increases beyond 76\% before experiencing an abrupt decline, surpassing the performance using CG alone. From this observation, we hypothesize that rapid attention shifting is most effective in shallower layers, where block effects have not yet blended with other image features, making them easier to isolate and manipulate.

\subsection{Visualization of Distributions and Features}
To illustrate how compression artifacts can mislead detectors and to verify the consistency of PLADA’s representations across different input qualities, we visualize the feature distributions of both the baseline and PLADA models using t-SNE in a 2D space. This visualization is performed on two datasets, each subjected to three compression types: Raw, Random Compression (Random), and Static Compression (Static). Random Compression  samples quantization parameters from the range $[30,100]$ for each image, while Static Compression applies a fixed quantization parameter of 50. The visualization results are shown in Figure \ref{feature visual}. When analyzing the baseline model's performance, it is evident that it can effectively distinguish between fake and real images on raw datasets. However, when images are compressed, the emergence of block effects introduces deceptive cues. Consequently, the features of compressed deepfakes tend to cluster closely with other features, indicating that the block effects successfully deceive the baseline model, leading to a decline in its performance. The similarity between the block effects and deepfake artifacts is likely a crucial factor in the reduced detection capabilities of the baseline model when handling compressed images. In contrast, PLADA exhibits remarkable resilience to compression. Across three types of data, it maintains the ability to distinguish between deepfake and real images while enhancing the separation between their features.

In summary, the proposed PLADA framework exhibits superior robustness across varying input qualities, consistently  delivering strong performance in detecting deepfakes on OSNs.

\subsection{Discussion}

\subsubsection{Unpaired Data Challenges}
In this study, we underscore the pivotal significance of unpaired data in bolstering defenses against compression attacks and introduce the ODA module to explore the distributional correlations between unpaired and paired data. However, it is important to note that our ODA serves as a straightforward implementation for aggregating unpaired data. Given that unpaired data constitutes a substantial portion of online deepfake multimedia content, including both images and videos, there is a pressing need to explore alternative and more effective methodologies that place more attention to unpaired data.

\subsubsection{Deceptive Features in Deepfakes}
Previous researches have primarily focused on developing methods or models to learn deepfake artifacts from diverse perspectives, such as the frequency and spatial domains. However, little attention has been given to the fact that data on OSNs is rarely raw experimental data but is often processed through multiple digital image techniques, such as beautification, extensive filtering, and compression. In this paper, we initially identify the block effect, a spatial manifestation of compression, as visually similar to deepfake artifacts, which we refer to as ``deceptive deepfake-like features". By automatically ignoring these features, we enhance the robustness of our detector. We firmly believe that numerous operations or processes can generate deceptive features that mislead models into making false judgments. Therefore, we encourage future research to conduct a deeper analysis of the influence of digital techniques along with proposing new methods.

\subsubsection{Multi-modal Supervision and Signal Integration}
To generate additional guidance signals, we employ multi-task learning, a fundamental yet not necessarily the most efficient method. Beyond the realm of deepfake detection, significant advancements have been achieved in large language models (LLMs), multi-modal large language models, and large vision models (LVMs), which have demonstrates remarkable capabilities across diverse tasks. To the best of our knowledge, there have been only limited efforts within the deepfake detection field to use supplementary information created by these foundational models. Given that numerous resource from OSNs are used to train these models, they inevitably learn intriguing details about both authentic and deepfake images prevalent on OSNs. We propose that researchers incorporate these potent foundational models into the domain of deepfake detection while ensuring that efficiency remains a priority.

\section{Conclusion}\label{Sec:conclusion}
This paper pioneers the study of the block effect, identifying it as a deceptive artifact that resembles deepfakes. To address this challenge with limited unpaired data, we introduce PLADA, a novel framework consisting of two core modules: ODA and B2E. PLADA effectively navigates the complexities posed by varying data qualities and compression methods. ODA leverages unpaired data to bridge the distribution gap between compressed and raw images through simple aggregation centers. Meanwhile, B2E captures guidance signals to redirect the model's attention to deepfake artifacts. Notably, PLADA enhances computational efficiency by eliminating the need for gradient correction. Extensive experiments across 26 widely used datasets under various test conditions demonstrate that PLADA significantly outperforms existing state-of-the-art models.

\bmhead{Funding Declaration}
This work was supported by National Natural Science Foundation of China (No.U24B20179, 62372451, 62192785), Beijing NSF (No.L242021, JQ24022).

\bmhead{Data Availability Statement}
All datasets and source code used in this study are publicly available. The access paths are as follows:
\textit{ForenSynths} dataset: \url{https://doi.org/10.1109/CVPR42600.2020.00872}, \textit{GANGen-Detection} dataset: \url{https://github.com/chuangchuangtan/GANGen-Detection}, \textit{Ojha-test dataset}: \url{https://doi.org/10.1109/CVPR52729.2023.02345},
\textit{DiffusionForensics} dataset: \url{https://doi.org/10.1109/CVPR52733.2024.02657}.
The implementation code is available at: \url{https://github.com/ManyiLee/PLADA}.
\bibliographystyle{plainnat}
\bibliography{sn-bibliography}

\end{document}